\documentclass[letterpaper, 10 pt, conference]{ieeeconf}  

\IEEEoverridecommandlockouts                              

\usepackage{graphics} 
\usepackage{epsfig} 
\usepackage{amsmath} 
\usepackage{lipsum}
\usepackage{verbatim}
\usepackage{caption}
\usepackage{url}
\usepackage{todonotes}
\usepackage[hidelinks]{hyperref}
\usepackage{subfigure}
\usepackage{caption}
\title{\LARGE \bf
RISAS: A Novel Rotation, Illumination, Scale Invariant\\ Appearance and Shape Feature
}

\author{Kanzhi Wu$^{1}$, Xiaoyang Li$^{2}$, Ravindra Ranasinghe$ ^{1} $, Gamini Dissanayake$^{1}$ and Yong Liu$^{2}$
\thanks{$^{1}$Kanzhi Wu, Gamini Dissanayake and Ravindra Ranasinghe are with Centre for Autonomous Systems, University of Technology Sydney, Australia, 2007 
        {\tt\small kanzhi.wu, gamini.dissanayake, ravindra.ranasinghe@uts.edu.au}}%
\thanks{$^{2}$Xiaoyang Li and Yong Liu are with Institute of Cyber-Systems and Control, Zhejiang University, China, 310027 {\tt\small zjulixiaoyang @163.com, yongliu@iipc .zju.edu.cn}}%
}

\begin{document}

\maketitle
\thispagestyle{empty}
\pagestyle{empty}

\begin{abstract}
This paper presents a novel appearance and shape feature, RISAS, which is robust to viewpoint, illumination, scale and rotation variations. RISAS consists of a keypoint detector and a feature descriptor both of which utilise texture and geometric information present in the appearance and shape channels. A novel response function based on the surface normals is used in combination with the Harris corner detector for selecting keypoints in the scene. A strategy that uses the depth information for scale estimation and background elimination is proposed to select the neighbourhood around the keypoints in order to build precise invariant descriptors. Proposed descriptor relies on the ordering of both grayscale intensity and shape information in the neighbourhood. Comprehensive experiments which confirm the effectiveness of the proposed RGB-D feature when compared with CSHOT \cite{tombari11cshot} and LOIND\cite{feng_icra_2015} are presented. Furthermore, we highlight the utility of incorporating texture and shape information in the design of \textit{both} the detector and the descriptor by demonstrating the enhanced performance of CSHOT and LOIND when combined with RISAS detector.
\end{abstract}

\section{INTRODUCTION}

Feature matching is a fundamental problem in both computer vision applications (e.g., object detection and image retrieval ) and robotic tasks (e.g., vision based Simultaneous Localisation and Mapping). Two critical steps toward finding robust and reliable correspondences are: 1) extracting discriminative keypoints, 2) building invariant descriptors. Over the past decades, there have been enormous progresses in developing robust features in two-dimensional image space such as SIFT(Scale Invariant Feature Transform) \cite{lowe_ijcv_2004}, SURF(Speed-Up Robust Feature) \cite{bay_eccv_2006} and ORB(Oriented FAST and Rotated BRIEF) \cite{rublee_iccv_2011}. These methods achieve excellent performance under significant scale and rotation variations when rich texture information is available, however, their performances dramatically degrade under illumination variations or in environments with poor texture information.

\begin{figure}[htbp]
	\centering
	\includegraphics[width=\linewidth]{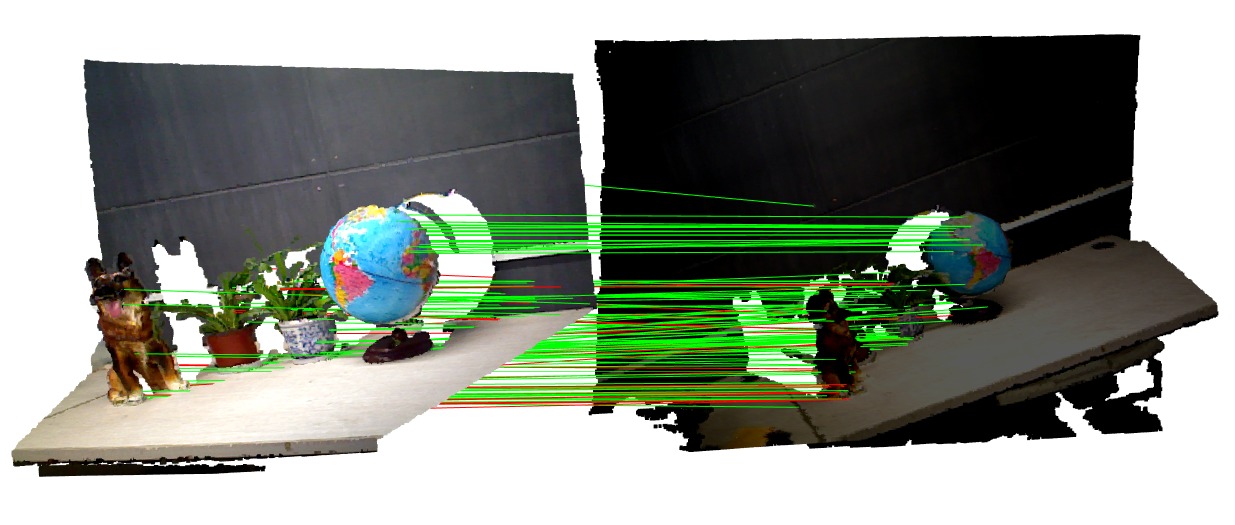}%
	\caption{Feature matching results using RISAS. The correct matches are shown using green lines and incorrect matches are denoted by red lines.}
	\vspace{-5mm}
	\label{fig::result_demon}
\end{figure}

With the development of low-cost, real-time depth sensors such as Kinect and Xtion, the geometric information of the environment can be accessed easily, thus it is now prudent to consider geometric information in building local descriptors. Spin Image\cite{johnson_pami_1999} is one of the well-known 3D descriptors which is widely used in 3D surface registration tasks. Rusu et al. \cite{rusu08persistentpoint}\cite{rusu_icra_2009}\cite{rusu2010vfh} also made tremendous contributions and proposed various depth descriptors such as PFH(Persistent Feature Histogram) and NARF(Normal Aligned Radial Feature). Despite the above developments, relying on depth information alone makes the correspondence problem more challenging due to two facts: 1) sensor information from Kinect and Xtion is generally incomplete 2) depth image is much less informative compared with RGB/grayscale image, particularly on regular shaped surfaces. Due to the complementary nature of RGB/grayscale information and depth information, combining appearance and geometric information is a promising direction to build descriptors and improve matching performance. CSHOT(Color Signature of Histogram and OrienTation) and BRAND(Binary Robust Appearance and Normal Descriptor) are examples of RGB-D descriptors. However, there is no specifically designed keypoints detector for these descriptors. Thus the selected keypoints may not reflect the best available regions in the scene for robust descriptor matching. 


The main contribution of this paper is a novel Rotation, Illumination and Scale invariant Appearance and Shape feature (RISAS) which tightly couples a discriminative RGB-D keypoint detector and an invariant feature descriptor. As a result of using texture and shape information in the design of both the detector and the descriptor, RISAS shows superior performance over current state-of-the-art methods under various conditions. Fig. \ref{fig::result_demon} demonstrates the capability of RISAS for obtaining correspondences under severe rotation, scale and illumination changes. Furthermore, benefits of using a RGB-D keypoint detector is highlighted by the enhanced performance of CSHOT descriptor when combined with the RISAS detector. In addition, we also make available a novel dataset which can be used for future evaluations of RGB-D detectors and descriptors.

The paper is structured as follows: Section II reviews the related work on both texture and shape based detectors and descriptors. In Section III, we introduce the proposed novel keypoint detector computed using both grayscale image and depth information. A novel RGB-D descriptor, that is built on LOIND(Local Ordinal Intensity and Normal Descriptor)\cite{feng_icra_2015} with significant enhancements is also described in Section III. In Section IV, the proposed feature (RISAS) is experimentally evaluated  using an existing public domain dataset as well as with a new dataset that includes variations in viewpoint, illumination, scale and rotation separately. Conclusion and future work are discussed in Section V.

\section{RELATED WORK}

In general, feature extraction can be separated into two sub-problems: keypoint detection and descriptor construction. Some of the feature extraction algorithms such as  SIFT\cite{lowe_ijcv_2004} and SURF\cite{bay_eccv_2006} tightly couple these two steps while methods such as FAST(Features from Accelerated Segment Test) and BRIEF(Binary Robust Independent Elementary Features) only focus on either keypoint detection or feature description.
\subsection{2D Appearance Features}

SIFT is one of the most well-known visual features \cite{lowe_ijcv_2004}. SIFT combines a Difference-of-Gaussian interest region detector and a gradient orientation histogram as the descriptor. By constructing the descriptor from a scale and orientation normalised image patch, SIFT exhibits robustness to scale and rotation variations.  SURF, proposed by Bay et al.\cite{bay_eccv_2006}, relies on integral images for image convolution. SURF uses a Hessian matrix-based measure for the detector and a distribution-based descriptor. Calonder et al.\cite{calonder2010brief} proposed BRIEF which uses a binary string as the descriptor. BRIEF feature takes relatively less memory and can be matched fast using Hamming distance in real-time with very limited computational resources. However, BRIEF is not designed to be robust to scale variations. Leutenegger et al.\cite{leutenegger2011brisk} proposed BRISK(Binary Robust Invariant Scalable Keypoints) which has a scale invariant keypoint detector and binary string like descriptor. ORB, another well-known binary feature, proposed by Rublee et al.\cite{rublee2011orb}, has been widely used in SLAM community\cite{raul_rss_15}. ORB is invariant to rotation variations and more robust to noise compared with BRIEF.

\subsection{3D Geometric Features}

In order to select salient keypoints from geometric information, researchers have adopted different criteria to evaluate the distinctiveness of the points in the scene, e.g.,  the normal vector of the surface and curvature of the mesh. Survey paper from Tombari et al. \cite{tombari2013performance} categorises 3D keypoint detectors into 2 classes: \textit{fixed-scale} detectors and \textit{adaptive-scale} detectors and provides a detailed comparison of existing 3D keypoint detectors. Hebert contributed several well-known \textit{adaptive-scale} detectors such as LBSS (Laplace-Beltrami Scale-Space) and MeshDoG\cite{unnikrishnan2008multi}. Zhong et al. proposed Intrinsic Shape Signature (ISS)\cite{zhong2009intrinsic} to characterise a local/semi-local region of a point cloud and ISS had been combined with various 3D descriptors in RGB-D descriptor evaluation\cite{guo2016comprehensive}. 

Descriptors can also be constructed using 3D geometric information. Johnson and Hebert \cite{johnson_pami_1999} proposed spin image which is a data level descriptor that can be used to match surfaces represented as meshes. With the development of low-cost RGB-D sensors, geometric information of the environment can be easily captured thus 3D shape descriptors have attracted renewed attention. More recent developments include PFH\cite{rusu08persistentpoint}, FPFH(Fast PFH) and SHOT(Signature of Histograms of OrienTations). Rusu et al. proposed PFH \cite{rusu08persistentpoint} which is a multi-dimensional histogram which characterises the local geometry of a given keypoint. PFH is invariant to position, orientation and point cloud density. Enhanced version of PFH, termed FPFH\cite{rusu_icra_2009} reduces the complexity of PFH from $ O(k^{2}) $ to $ O(k) $ where $ k $ is the number of points in the neighbourhood of the keypoint. SHOT descriptor proposed by Tombari et al.\cite{tombari2010} is another example of a widely used local surface descriptor. SHOT encodes the histograms of the surface normals in different partitions in the support region. 

Despite the progress in 3D shape descriptors, because of the fact that 3D geometric information is not sufficiently rich compared with the RGB or grayscale image, a shape descriptor alone is unable to provide reliable and robust feature matching results.

\subsection{Combined Appearance and Depth Features}

Lai et al. \cite{lai11} have demonstrated that by combining RGB and depth channels together, better object recognition performance can be achieved. Tombari et al. \cite{tombari11cshot} developed CSHOT via incorporating RGB information into original SHOT descriptor. Nascimento et al. \cite{nascimento_iros_2012} proposed a binary RGB-D descriptor BRAND which encodes local information as a binary string thus makes it feasible to achieve low memory consumption. They have also demonstrated the rotation and scale invariance of BRAND. More recently, Feng et al. \cite{feng_icra_2015} proposed LOIND which encodes the texture and depth information into one descriptor supported by orders of intensities and angles between normal vectors.

Most of the current RGB-D fused descriptors adopt traditional 2D keypoint detectors that rely only on appearance information. For instance, BRAND\cite{nascimento_iros_2012} is combined with CenSurE(Centre Surround Extremas\cite{agrawal2008censure}) detector and LOIND\cite{feng_icra_2015} uses keypoints from multi-scale Harris detector. In CSHOT\cite{tombari11cshot}, in order to eliminate the influence of detector, the keypoints are selected randomly from the model. Clearly, selecting keypoints by exploiting geometrically information-rich regions in the scene has the potential to enhance the matching performance of a RGB-D descriptor. In this work, we propose a keypoint detector and descriptor which relies on information from both appearance and depth channels. It is demonstrated that using both texture and depth information leads to a detector which will extract keypoints that are more distinctive in the context of a descriptor that also uses similar information, thus improving the discriminativeness of the descriptor. 

\section{METHOD}
In this section, we describe the proposed Rotation, Illumination and Scale invariant Appearance and Shape feature, RISAS, in detail. RISAS is built on our previous work \cite{feng_icra_2015}.  The detector and descriptor are explained in detail in Section.\ref{method::detector} and Section.\ref{method::descriptor}.

\subsection{Keypoint Detector}\label{method::detector}

The main advantage of using depth information in keypoint detection is the fact that information rich regions in the depth channel are also given due consideration without being ignored when these regions lack texture information.Both the proposed detector and the descriptor use similar information and thus are tightly coupled giving rise to superior matching performance.

\begin{figure}[htbp]
	\centering
	\includegraphics[width=0.9\linewidth]{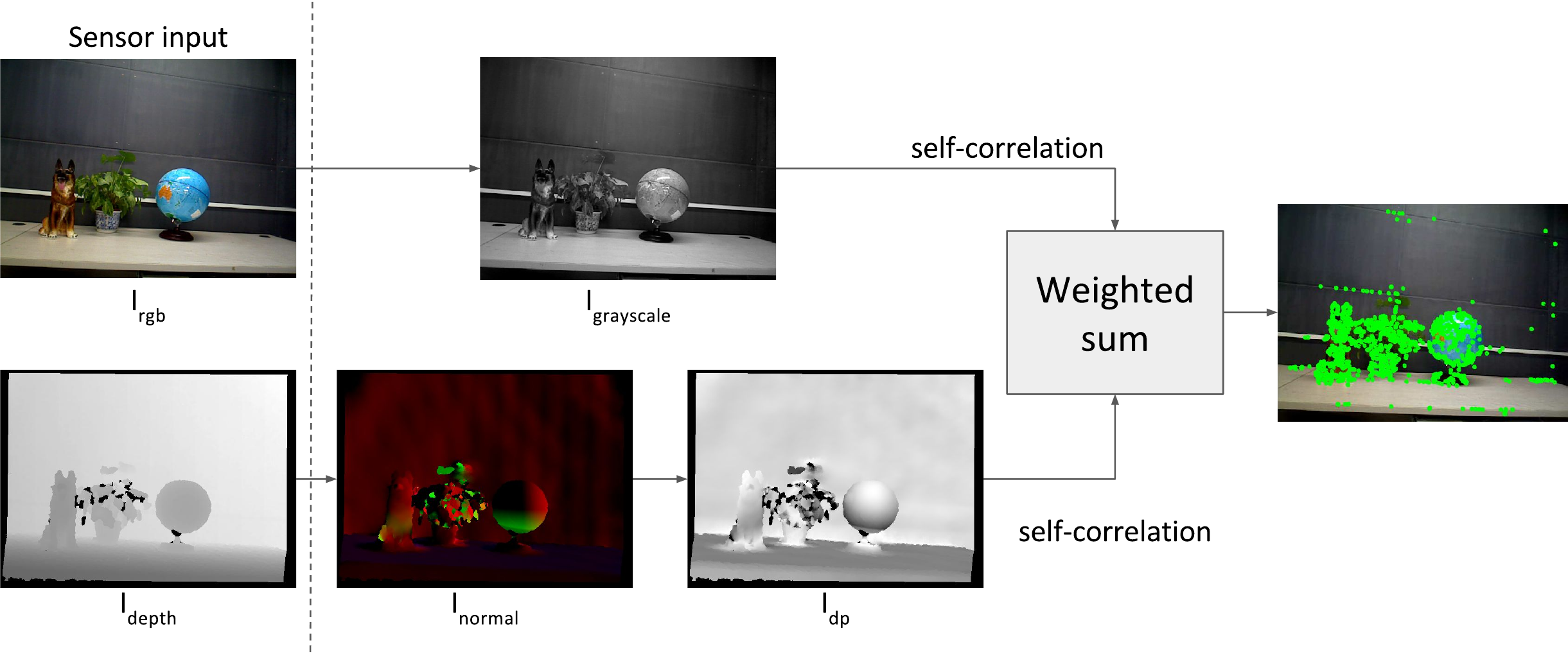}%
	\caption{Flowchart of the proposed RGB-D keypoint detector. $ I_{rgb} $ is the original RGB image and $ I_{grayscale} $ is the converted grayscale image. $ I_{normal} $ is the $ 3 $ channel normal vector image and $ I_{dp} $ is the dot product image.}
	\label{fig::flowchart_detector}
\end{figure}

The flowchart of the keypoint detection method is shown in Fig. \ref{fig::flowchart_detector} and the key steps are listed below:

\begin{enumerate}
	\item For each point in the depth image $ I_{depth} $, we calculate the surface normal vector. From the three components of the normal vector, we create the corresponding normal image $ I_{normal} $ with three channels. 
	
	\item Using $ I_{normal} $, we compute the three angles $ [\alpha, \beta, \gamma] $ between each normal vector and the $ [x, y, z] $ axis of the camera coordinate system respectively. The angle range $ [0, \pi] $ is segmented into $ n_{s} $ sectors labelled with $ [1, ..., n_{s}] $ and each computed angle is mapped into one of these sectors. In this work, $ n_{s} $ is set to be $ 4 $ as shown in Fig. \ref{fig::mainNormal}. For example, normal vector $ \mathbf{n} = \left[ \frac{\sqrt{3}}{3},\frac{\sqrt{3}}{3},\frac{\sqrt{3}}{3} \right] $ has the $ \left[ \alpha, \beta, \gamma \right]  = \left[ 54.7^{\circ}, 54.7^{\circ}, 54.7^{\circ} \right] $ will be labelled as $ \left[2, 2, 2\right] $;
	
	\begin{figure}[htbp]
		\centering
		\includegraphics[width=0.7\linewidth]{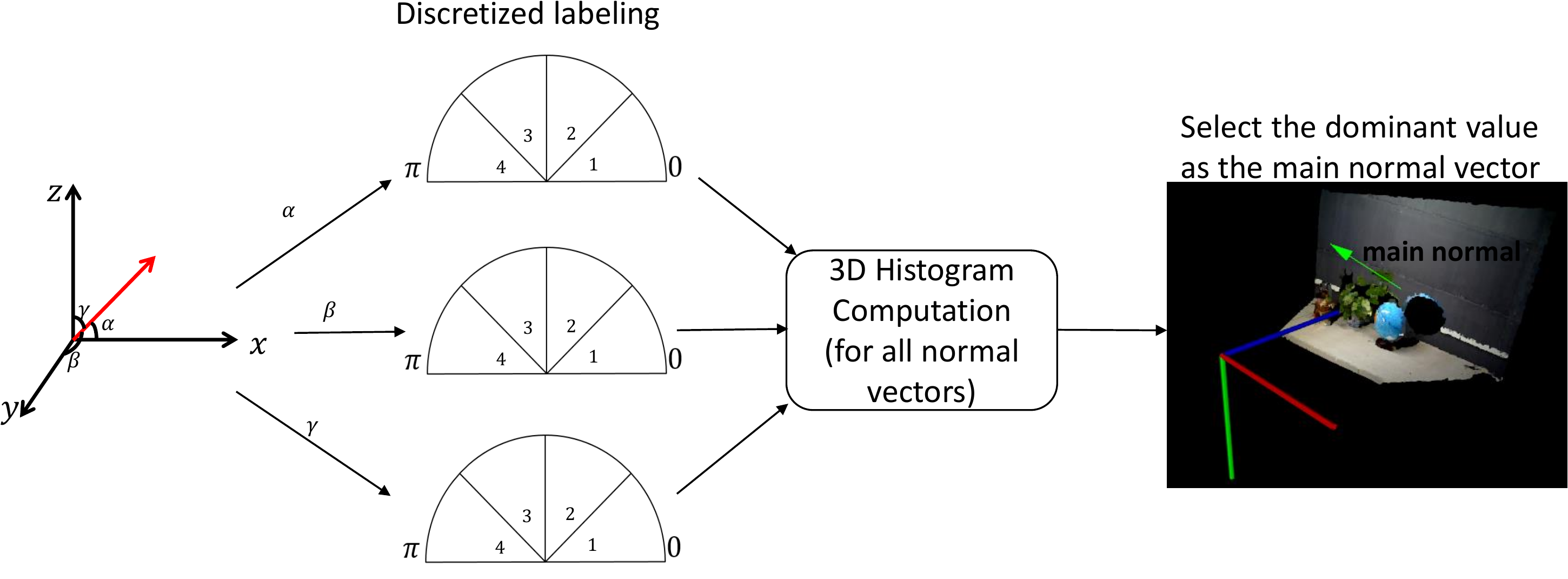}%
		\caption{Flowchart of calculating the main normal.}
		\label{fig::mainNormal}
	\end{figure}
	
	\item Using this labelled image, we build a statistical histogram to capture the distribution of labels along each channel. From this histogram, we choose the highest entry for each channel and use the corresponding label $ [n_{X}, n_{Y}, n_{Z}] $ to represent the most frequent label where $ n_{X}, n_{Y}, n_{Z} \in \{1, ..., n_{s}\} $. Using these three values, we define the ``main'' normal vector $ \mathbf{n}_{main} $ of the depth image $ I_{depth} $. 
	
	
	\item Calculate the dot-product between $ \mathbf{n}_{main} $ and each normal vector in $ I_{normal} $. This describes the variation of information in depth channel. We then normalise the dot product value into range $ [0,255] $. Using this value, we create the novel dot-product image $ I_{dp} $ which is approximately invariant to the viewpoint of the sensor. 
	
	\item We adopt the similar principle as in Harris detector to compute the response value $ E(u,v) $ using the grayscale image $ I_{grayscale} $ and the dot product image $ I_{dp} $. The response value is thresholded to select points that show an extreme value in the weighted sum of two response values from $ I_{grayscale} $ and $ I_{dp} $, as shown in Eq.\ref{eq::autocorrelation}:
	\begin{equation}\label{eq::autocorrelation}
		\begin{aligned}
		E\left(u, v\right) & = \sum_{x,y}\omega(x, y) [ \tau \left( I(x+u, y+v) - I(x,y) \right)^{2}\\
		& + (1-\tau)\left( P(x+u, y+v) - P(x,y) \right)^{2}]
		\end{aligned}
	\end{equation}
	where $ \left(u, v\right) $ is the keypoint coordinate in image space and $ \omega(x, y) $ is the window function centred at $ (u, v) $ which is a Gaussian function in this paper. $ I(u,v) $ is the intensity value at $ (u, v) $ and $ P(u, v) $ is the normalized dot product value at $ (u, v) $. Empirical study shows that $ \tau $ plays a critical role in balancing appearance information and geometric information in keypoint detection. Because of the fact that rgb/grayscale image is more information rich compared with depth image and provides more variations, $ \tau $ should assign larger value to rgb image. Fig. \ref{fig::tau-ratio} provides precision-recall curves for different $ \tau $ value for the same scenario. We selected $ \tau = 0.8 $ after numerical experimental evaluations.
	\begin{figure}[htbp]
		\centering
		\includegraphics[width=0.65\linewidth]{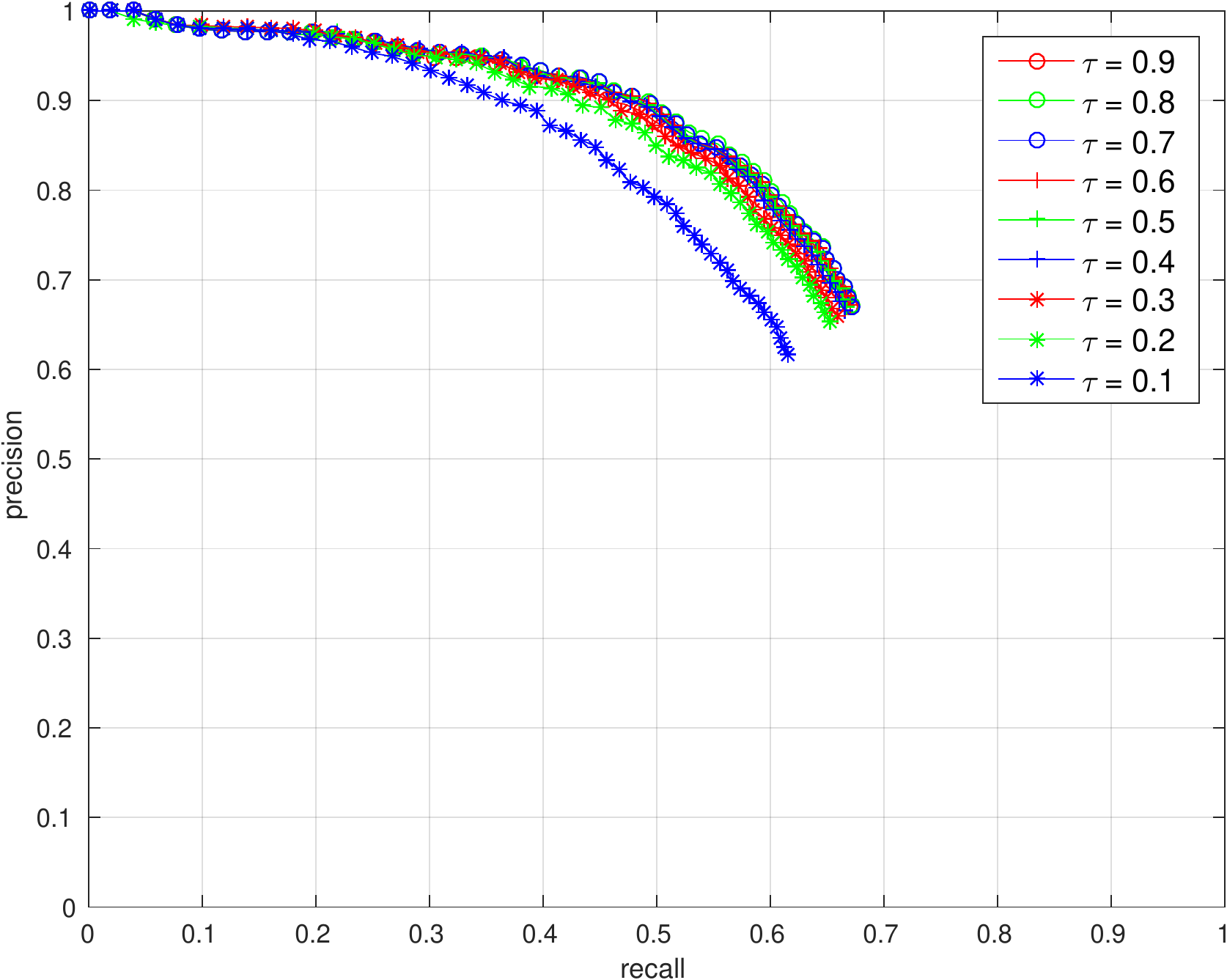}%
		\caption{Precision-Recall curves for difference $ \tau $ value.}
		\label{fig::tau-ratio}
		\vspace{-2mm}
	\end{figure}
\end{enumerate}
This strategy clearly identifies keypoints from regions that are information rich in both appearance and geometry.

\subsection{Feature Descriptor}\label{method::descriptor}

\subsubsection{Scale Estimation and Neighbourhood Region Selection}

For grayscale images, the scale of the keypoint is estimated by finding the extreme value in scale space using image pyramid.  Typical examples are as SIFT\cite{lowe_ijcv_2004} and SURF\cite{bay_eccv_2006}. With the development of modern RGB-D sensors such as Kinect and Xtion, the scale can be easily measured using the depth information captured from the sensor. In both LOIND\cite{feng_icra_2015} and BRAND\cite{nascimento_iros_2012}, the following empirical equation scales the distance range between $ \left[ 2, 8 \right] $ into scale range $ \left[ 1, 0.2 \right] $ in a linear relationship. Scale value for distance less than $ 2 $m is truncated as $ 1 $.
\begin{equation}
s = \max\left( 0.2, \frac{3.8-0.4\max(2, d)}{3}\right)
\label{eq::scale_est}
\end{equation}
After $ s $ is estimated, the neighbourhood region that is used to build the descriptor is selected with radius $ R $ in a linear relationship with scale value $ s $, as shown in \cite{feng_icra_2015}\cite{nascimento_iros_2012}. A critical deficiency in their approach is that the neighbourhood region is selected without considering the geometric continuity. In the following we present a more accurate method for selecting the neighbourhood region from which the descriptor is built.

\begin{figure}[htbp]
	\centering
	\includegraphics[width=0.8\linewidth]{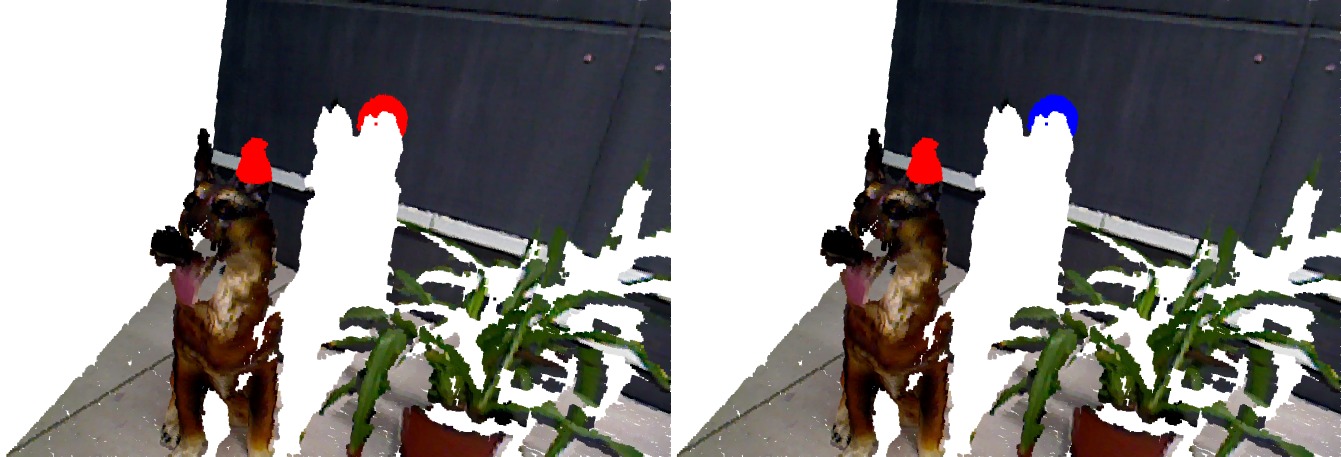}%
	\caption{Neighbourhood selection: The default strategy (left) selects the whole region (shown in red) which covers both foreground and background area. However, the introduced background points have an adverse effect on the local descriptor. Our approach (right) eliminates the background points (shown in blue) and constructs the descriptor using the foreground (shown in red) only, leading to more robust descriptor matching performances.}
	\label{fig::neighbourhood_selection}
\end{figure}

\begin{enumerate}
	\item Based on Eq. \ref{eq::scale_est}, initial value of the scale $s$ is estimated. The radius $R$ of the patch is computed using Eq. \ref{eq::initalRadius} which was derived using extensive experimentation. 
	\begin{equation}\label{eq::initalRadius}
	R = \left( -5+25*\min \left( 3,\frac{\max(0.2.s_{max})}{\max(0.2.s_{min})}\right) \right)\cdot s
	\end{equation}
	Where $ s_{max} $ and $ s_{min} $ are the maximum and minimum scale values in the image. 
	It is an empirical value based on the experiments, if scale varies gently in the neighbourhood region, we can choose a smaller $ R $ and vice versa. We denote the patch centred at keypoint $ k_{i} $ in 2D image space as $ \mathbf{P}^{uv}(k_{i}) $ and the corresponding patch in 3D point cloud space is represented as $ \mathbf{P}^{xyz}(k_{i}) $;
	
	\item For each point $p \in \mathbf{P}^{xyz}(k_i)$, we remove the outlier neighbouring points from the keypoint $ k_{i} $ according to Eq.\ref{eq::ifInlier}. This step of eliminating the background was found to produce significant improvements in the matching performance.
	
	\begin{equation}
	f(p)=
	\begin{cases}
	1& \text{ if } 
	\begin{Vmatrix}
	p-k_i
	\end{Vmatrix}<t \\ 
	0& \text{ otherwise }   
	\end{cases}
	\label{eq::ifInlier}
	\end{equation}
	
	where $t$ is the threshold and set to be $ 0.1 $ meter in this work. We only keep the neighbouring points with $ f(p) = 1 $;
	
	\item We conduct ellipsoid fitting for the processed 3D neighbouring points $\bar{ \mathbf{P} }^{xyz} (k_i)$ based on the following equation.
	\begin{equation}
	\frac{(x-x_{k_{i}})^2}{a^2}+\frac{(y-y_{k_{i}})^2}{b^2}+\frac{(x-z_{k_{i}})^2}{c^2} = 1
	\end{equation}
	where $ a, b $ and $ c $ are the length of the axes. We project the 3D ellipsoid into the image space for the new accurate patch $\mathbf{\bar{P}}^{uv}$ with radius $ \bar{R} $ for further descriptor construction. 
\end{enumerate}

\subsubsection{Orientation Estimation}\label{method::descriptor::ori}

In LOIND\cite{feng_icra_2015}, the dominant orientation $ \theta $ of the selected patch is computed from the depth information only.  Although it works reasonably well under different scenarios it is sensitive to the noise in neighbourhoods where the normal vectors are similar to each other. In the following, we propose an alternative novel dominant orientation estimation algorithm which is more robust and efficient compared with LOIND\cite{feng_icra_2015}:

\begin{enumerate}
	
	\item Given the processed 2D patch $ \bar{ \mathbf{P} }^{uv} $ and 3D patch $ \bar{ \mathbf{P} }^{xyz} $, we adopt PCA to compute the eigenvalues $ \left[ e_{1}, e_{2}, e_{3} \right] $ (in \textit{descending} order) and corresponding eigenvectors $ \left[ \mathbf{v}_{1}, \mathbf{v}_{2}, \mathbf{v}_{3} \right] $.

	\item Given the eigenvectors $ \left[ \mathbf{v}_{1}, \mathbf{v}_{2}, \mathbf{v}_{3} \right] $, the 3D dominant orientation $ d_{3D} $ of the patch is computed as follows:
	
	\begin{equation}
	d_{3D}=
	\begin{cases}
	\frac{\mathbf{v}_1 \times \mathbf{v}_2}{|\mathbf{v}_1 \times \mathbf{v}_2 | } & \text{ if } (e_2 > \gamma e_{1}) \wedge (e_3 \le \gamma e_{1})  \\ 
	\text{rejected} & \text{ if } (e_2 > \gamma e_{1}) \wedge (e_3 > \gamma e_{1})\\
	\mathbf{v}_1 & \text{ others }   
	\end{cases}
	\label{eq::orientation}
	\end{equation}
	
	where $ \gamma $ is set within $ [0, 1] $. If the $ e_{1} $ is significantly larger than other two, the 3D dominant orientation is set to be the corresponding eigenvector $ \mathbf{v}_{1} $. If $ e_{2} $ is close to $ e_{1} $, both eigenvector $ \mathbf{v}_{1} $ and $ \mathbf{v}_{2} $ are considered in computing the dominant orientation by taking the cross-product of these two vectors. Further if both $ e_{3} $ and $ e_{2} $ are closer to $ e_{1}$ which means no clear differences between $ 3 $ eigenvalues, this keypoint is rejected because the depth channel won't be able to provide distinctive information. Threshold $ \gamma $ determines when the second eigenvalues $ e_{2} $ can be regarded as ``close" enough to the largest eigenvalue $ e_{1} $ which is set to be $ 0.8 $ through experiments.
	
	\item Project the 3D dominant direction $ d_{3D} $ into the image plane and get the 2D dominant direction $ d_{2D} $. We use $ \theta $ to denote the angle between $ d_{2D}$ and $ u $ axis in image space.
	
\end{enumerate}

\subsubsection{Descriptor Construction}\label{method::descriptor::descr}

\begin{figure}[htbp]
	\centering
	\includegraphics[width=\linewidth]{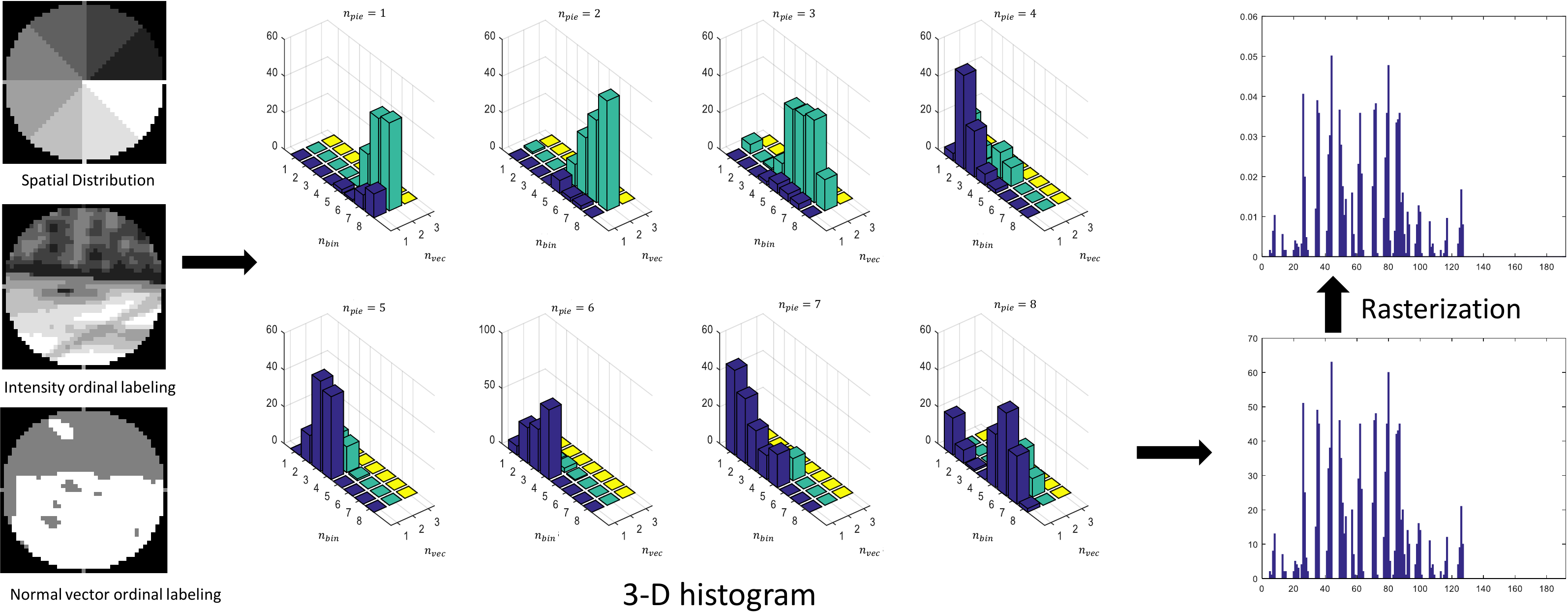}%
	\caption{Flowchart of the  RGB-D descriptor.}
	\label{fig::flowchart_descriptor}
\end{figure}

Based on the results from the above steps, we can construct the descriptor of keypoint $ k_{i} = \left[u, v\right] $ using the neighbourhood region with radius $R$ and the angle $\theta$. We follow the main ideas used in LOIND\cite{feng_icra_2015}. The descriptor is based on the relative order information in both grayscale and depth channels. The descriptor is constructed in a three-dimensional space, as show in Fig. \ref{fig::flowchart_descriptor} below where $ \left[x, y, z\right] $ axes denote the spatial labelling, the intensity labelling and the angles labelling respectively. 

\begin{enumerate}
	
	\item[-] \textbf{Encoding Spatial Distribution} \\For spatial distribution, the pixels in the region $(u,v,R,\theta)$ are labeled based on $n_{pie}$ equal-size spatial sectors. Larger the number of sectors, the more discriminative the descriptor, but this clearly effects on timing for both construction and matching. 

	\item[-] \textbf{Encoding Grayscale Information} \\Instead of constructing the descriptor in the absolute intensity space, we build the statistical histogram using the relative intensity with respect to the intensity value of the keypoint, in order to enhance illumination invariance. According to the rank of all the pixels in the patch,we group the intensity values into $n_{bin}$ equally sized bins. For example, given $ 100 $ intensity levels and $ 10 $ bins, eachbin has $ 10 $ intensity levels (i.e., orderings of $ [1,10], [11, 20], \dots, [91, 100] $ ).
	
	\item[-] \textbf{Encoding Geometrical Information} \\ Given the normal vector of each point, we first compute the dot product between the normal vector of the selected keypoint $ \mathbf{n}_{p_{k}} $ and the normal vector of each point in the neighbourhood patch $ \mathbf{n}_{p_{i}} $.
	
	\begin{equation}
	\rho_i=|\langle \mathbf{n}_{p_{k}}, \mathbf{n}_{p_{i}} \rangle| 
	\label{eq::dot_product}
	\end{equation}
	
	Due to the fact that normal vectors from small patches are similar to each other, thus the distribution of $ \rho_{i} $ is highly unbalanced where the majority of $ \rho_{i} $ falling into the range close to $ 1 $. We set a threshold $ \bar{\rho} = 0.9$ and any $ \rho_{i} \ge \bar{\rho} $ are grouped in to one category. The remaining dot products are ranked and grouped into $ n_{vec} $ bins.  Points are then labelled based on the group they belong to respectively. Therefore, in normal vector space encoding, there are overall $ n_{vec}+1 $ labels. 
	
\end{enumerate}

\begin{figure}[htbp]
	\centering
	\includegraphics[width=0.65\linewidth]{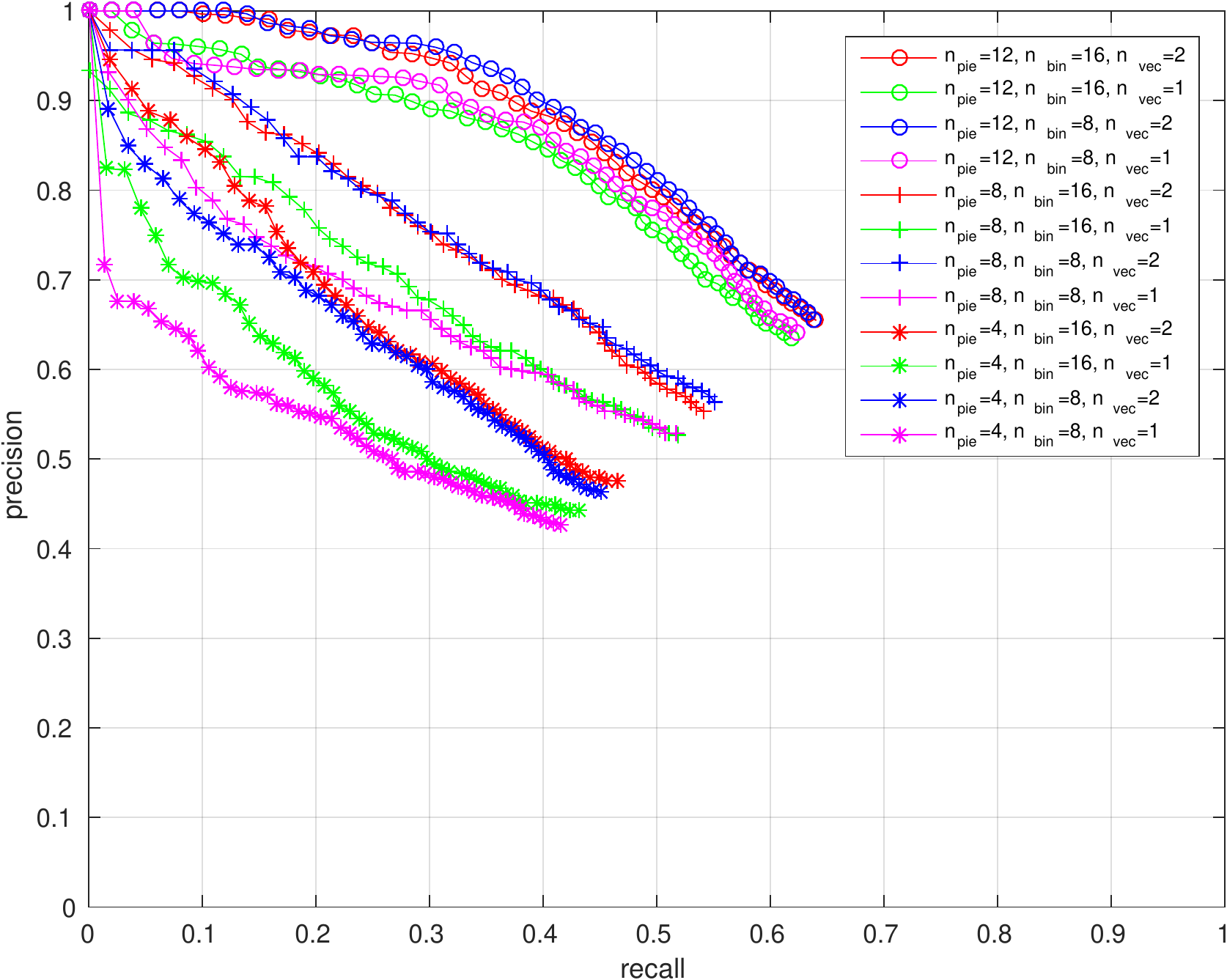}%
	\caption{Parameter selection for descriptor construction.}
	\label{fig::param-selection}
	\vspace{-5mm}
\end{figure}

During the empirical study, we tested $ 12 $ different combination of parameters $ n_{pie} = \left\lbrace4, 8, 12\right\rbrace, n_{bin} = \left\lbrace8, 16\right\rbrace $ and $ n_{vec} = \left\lbrace 1, 2\right\rbrace $. The precision-recall curves are presented in Fig. \ref{fig::param-selection}. Considering both performance and efficiency, in the experiments section, we set parameters as $ n_{pie} = 8,n_{bin} = 8, n_{vec} = 2  $ and we have a  192-dimensional ( $\dim = n_{pie}\cdot n_{bin}\cdot (n_{vec}+1) $) descriptor.

\section{Experimental Results}\label{chap::exp}

In this section, we compare the performance of RISAS against CSHOT, LOIND and other methods.  We also report the results obtained using SIFT, to highlight the value of using both appearance and depth channels. We use a public RGB-D dataset which is originally designed for object detection\footnote{\url{http://rgbd-dataset.cs.washington.edu/}}. This dataset does not include examples of rotation, scale or illumination changes independently and therefore is not able to fully illustrate the effectiveness of the RISAS in such situations. Therefore we have designed our own dataset for further detailed evaluations\footnote{This dataset can be downloaded from \url{http://kanzhi.me/rgbd-descriptor-dataset/} to make it possible for the community to use this in future research}. 

\subsection{Evaluation Method}

Firstly, we extract keypoints from two frames and construct the descriptors for all these keypoints. Nearest Neighbour Distance Ratio (NNDR) is used to establish the correspondences of keypoints between a pair of images. We use the reprojection error to determine whether a correspondence is correct using the equation below:
\begin{equation}\label{eq::correct}
|| p_{i} - \left( \mathbf{R}p_{j}+\mathbf{t} \right) || \le d_{\min}
\end{equation}
where $ p_{i} $ and $ p_{j} $ are 3d points from frames $ i $ and $ j $. $ \mathbf{R} $ and $ \mathbf{t} $ denote the groundtruth rotation and translation and are given during the evaluation. If the re-projection error is less than $ d_{\min} $(set to be $ 0.05 $ m), the match is regarded as a correct one. In the next subsection, we use the \textit{percentage of inliers} to describe the invariance of the features w.r.t scale variations and we adopt \textit{Precision-Recall} curves to evaluate the performance of the RGB-D features under other types of variations similar as \cite{mikolajczyk2005performance} .

\subsection{Experimental Results and Analysis}

In this section, we present the following  comparative experiments against our proposed RISAS feature:

\begin{enumerate}
	\item 3D ISS keypoint detector and RGB-D CSHOT descriptor: ISS has been combined with different 3D descriptors for evaluation in Guo et al.'s survey\cite{guo2016comprehensive}. Implementations of these in PCL\cite{rusu20113d} were used in our experiments. 
	\item Uniform sampled keypoints and RGB-D CSHOT descriptor: Uniform sampling method for keypoint detection was used in Aldoma et al.'s work \cite{aldoma2012tutorial} for 3D object recognition\footnote{Random sampling is used in the SHOT\cite{tombari2010} paper and CSHOT paper\cite{tombari11cshot}.} In our experiments, the uniform sampling method was adopted and the methods provided in PCL were used. 
	\item 2D SIFT keypoint detector and RGB-D CSHOT descriptor: We used publically available implementations of SIFT detector from VLFeat\cite{vedaldi08vlfeat} and CSHOT descriptor from PCL\cite{rusu20113d}. This was used as an example of combination between a 2D keypoint detector and a RGB-D descriptor.
	\item Proposed RISAS keypoint detector and RGB-D CSHOT descriptor: Matlab implementation of the RISAS detector together with the PCL implementation of CSHOT was used. 
	\item 2D SIFT feature (detector and descriptor) as implemented in VLFeat.
	\item Proposed RISAS keypoint detector and LOIND descriptor that were implemented in Matlab.
\end{enumerate} 

All of the experiments were performed on a standard desktop PC equipped with an Intel i5-2400 CPU. 

\subsection{Object Recognition Dataset}
\begin{figure}[!htbp]
	\centering
	\subfigure[Image 33 and 38.]
	{\includegraphics[width=0.62\linewidth]{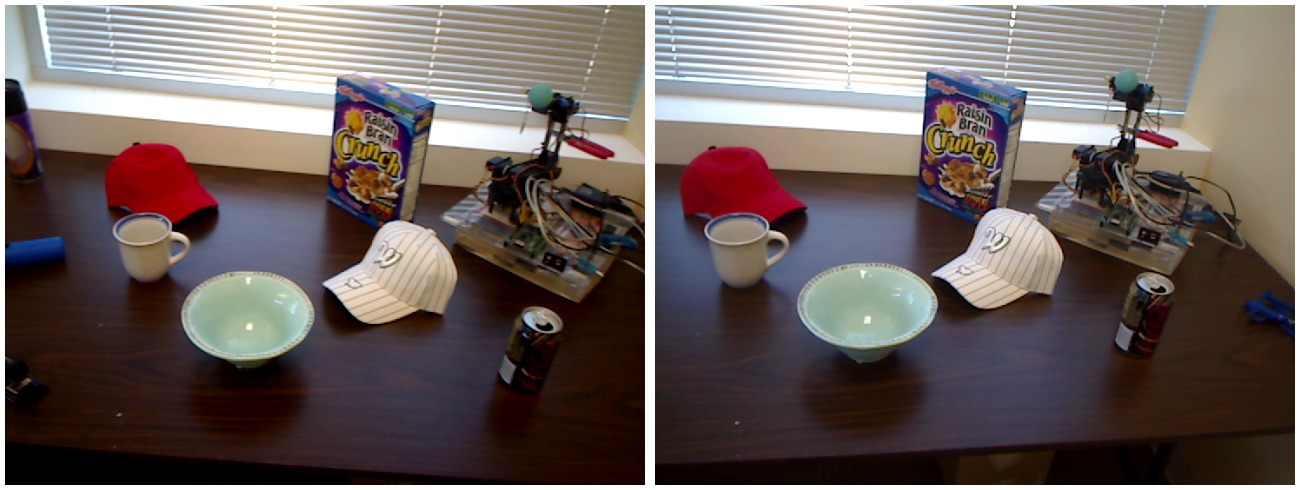}
		\label{fig:pub_10_20}}
	\subfigure[]
	{\includegraphics[width=0.32\linewidth]{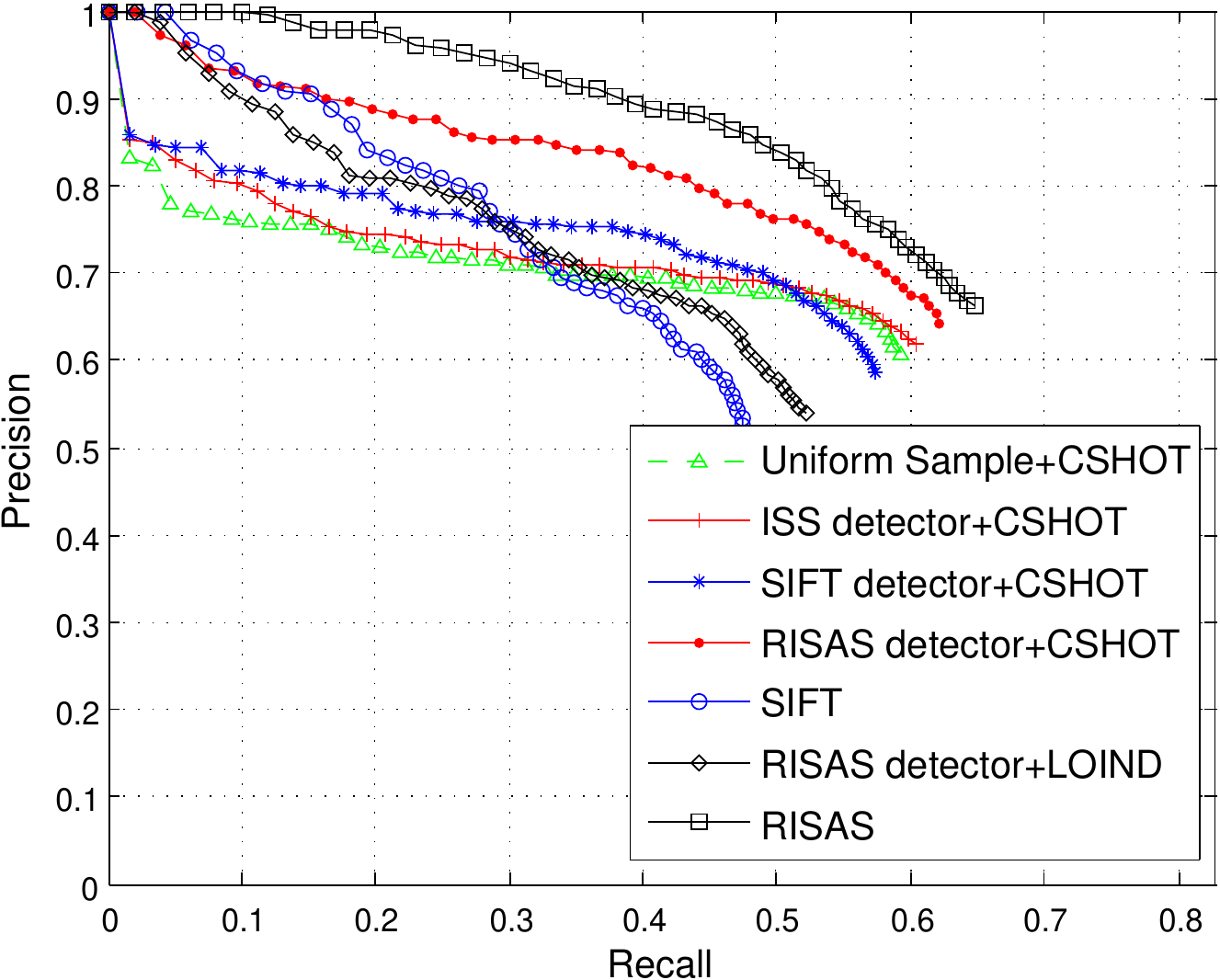}
		\label{fig:pub_10_20_pr}}
	\\
	\subfigure[Image 25 and 32.]
	{\includegraphics[width=0.62\linewidth]{./figures/newresults/rgb_public_1.jpg}
		\label{fig:pub_70_80}}
	\subfigure[]
	{\includegraphics[width=0.32\linewidth]{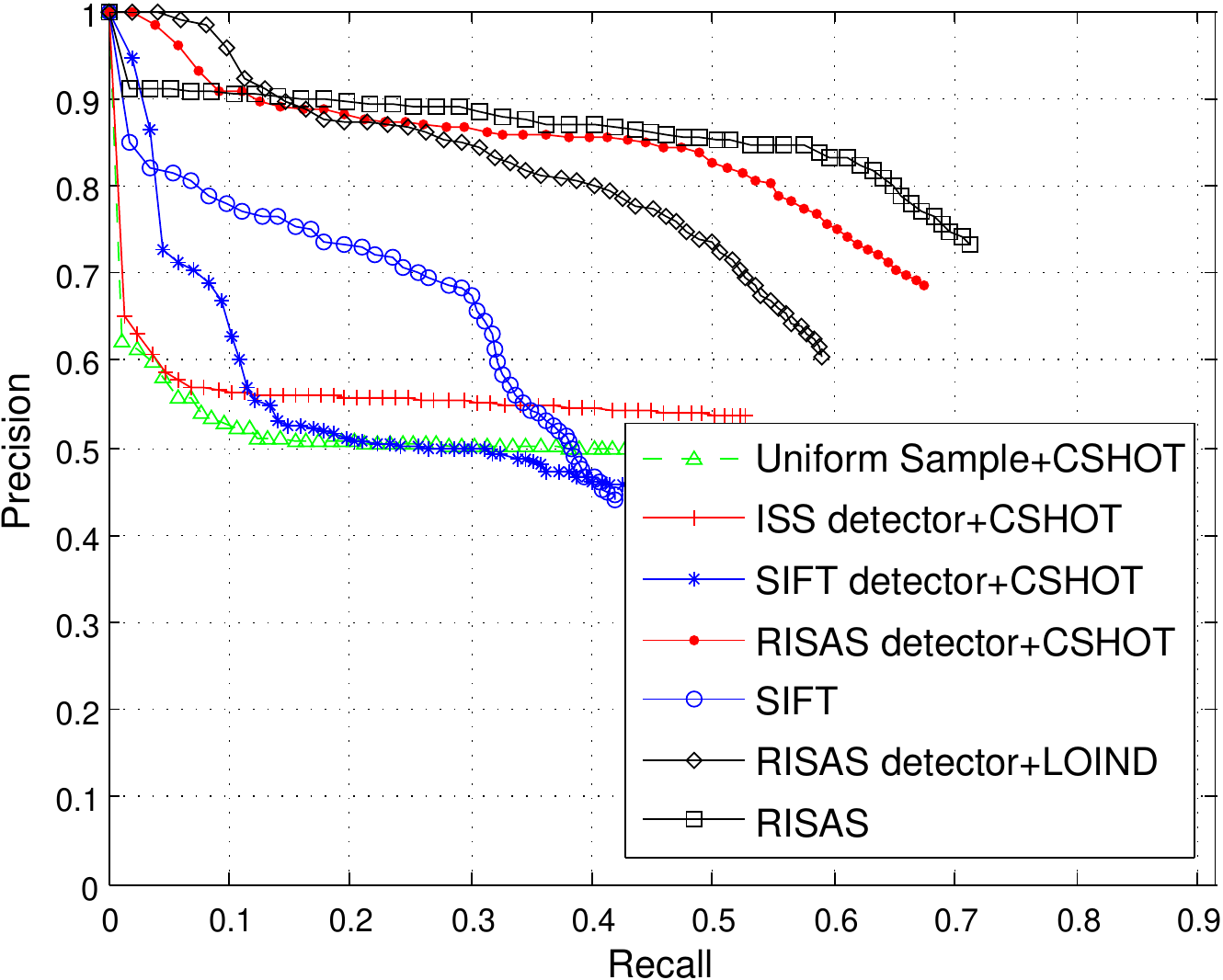}
		\label{fig:pub_70_80_pr}}	
	\caption{Evaluation results on RGB-D scene dataset.}\label{fig:pub_pr}
\end{figure}

We selected the information-rich sequence $ table\_1 $  from the RGB-D scene dataset \cite{lai11} and we present some of the results in Fig. \ref{fig:pub_pr}. As the figure indicates, RISAS and the combination of RISAS detector and CSHOT descriptor show larger area under the curve thus demonstrate the best performance.

\subsection{RGB-D Feature Evaluation Dataset}
In the constructed dataset, we consider four common variations: 1)  viewpoint, 2) illumination, 3) scale and 4) rotation.

\subsubsection{Viewpoint Invariance}   
\begin{figure}[htbp]
	\centering
	\subfigure[Image 12]
	{\includegraphics[width=0.42\linewidth]{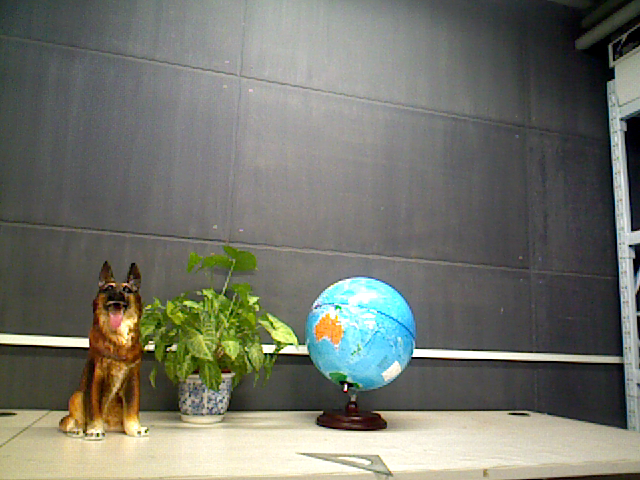}
		\label{fig:scale_1}}
	\subfigure[Image 24]
	{\includegraphics[width=0.42\linewidth]{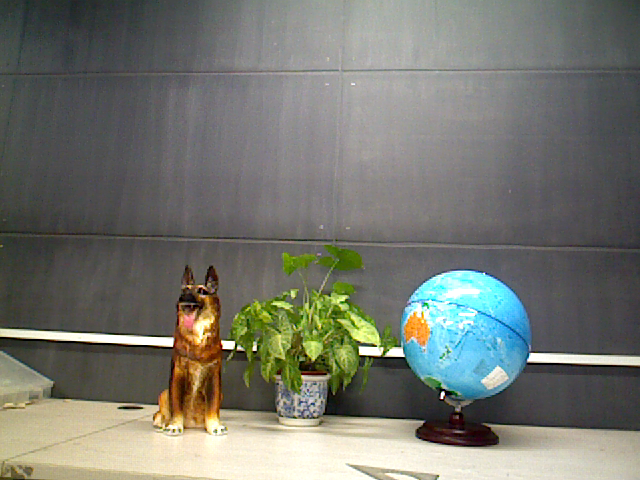}
		\label{fig:scale_10}}
	\caption{Example images of viewpoint variations.}\label{fig:vp_example}
	\vspace{-2mm}
\end{figure}

\begin{figure}[htbp]
	\centering
	\subfigure[Between image 12 and 1]
	{\includegraphics[width=0.42\linewidth]{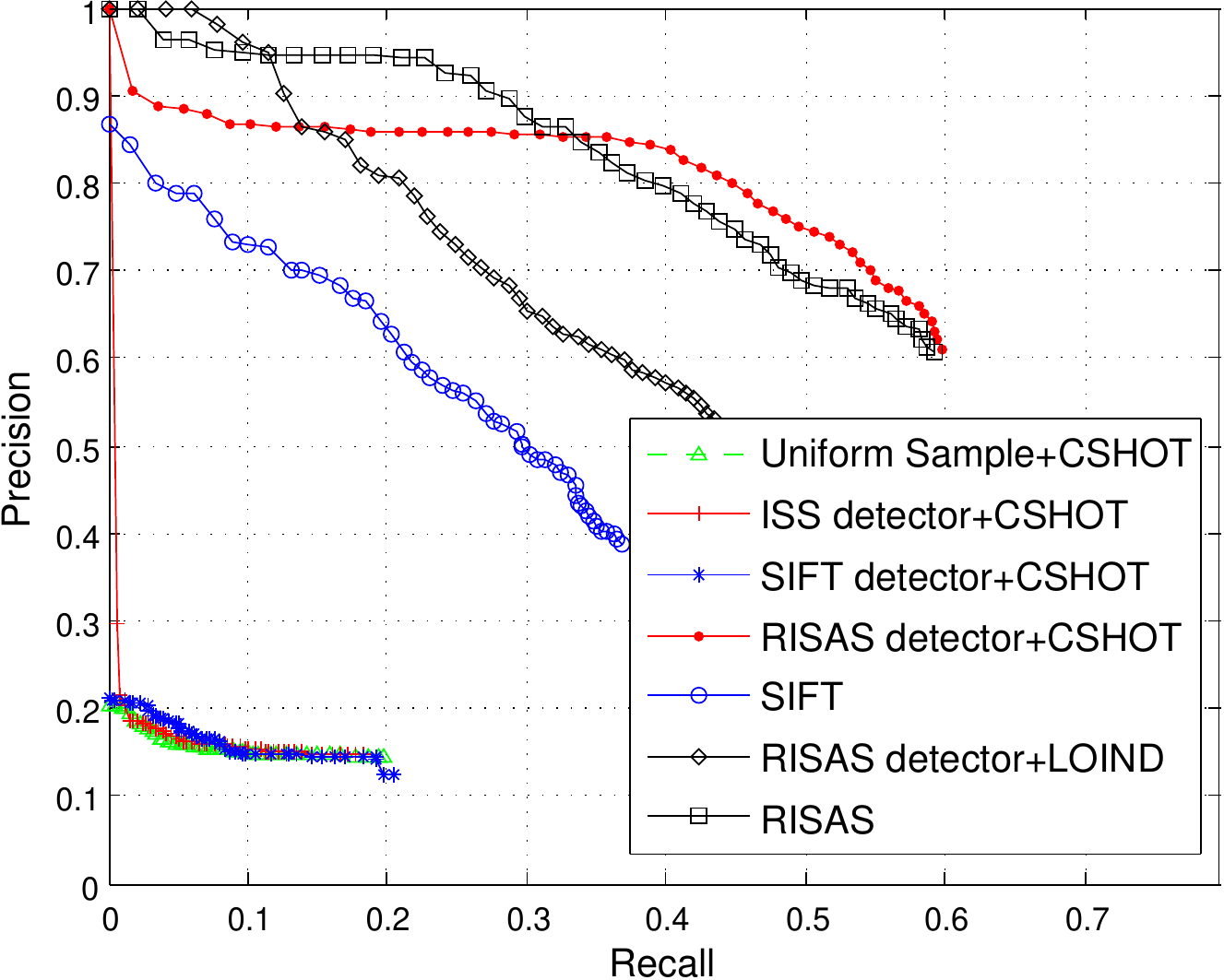}
		\label{fig:vc_pr1}}
	\subfigure[Between image 12 and 6]
	{\includegraphics[width=0.42\linewidth]{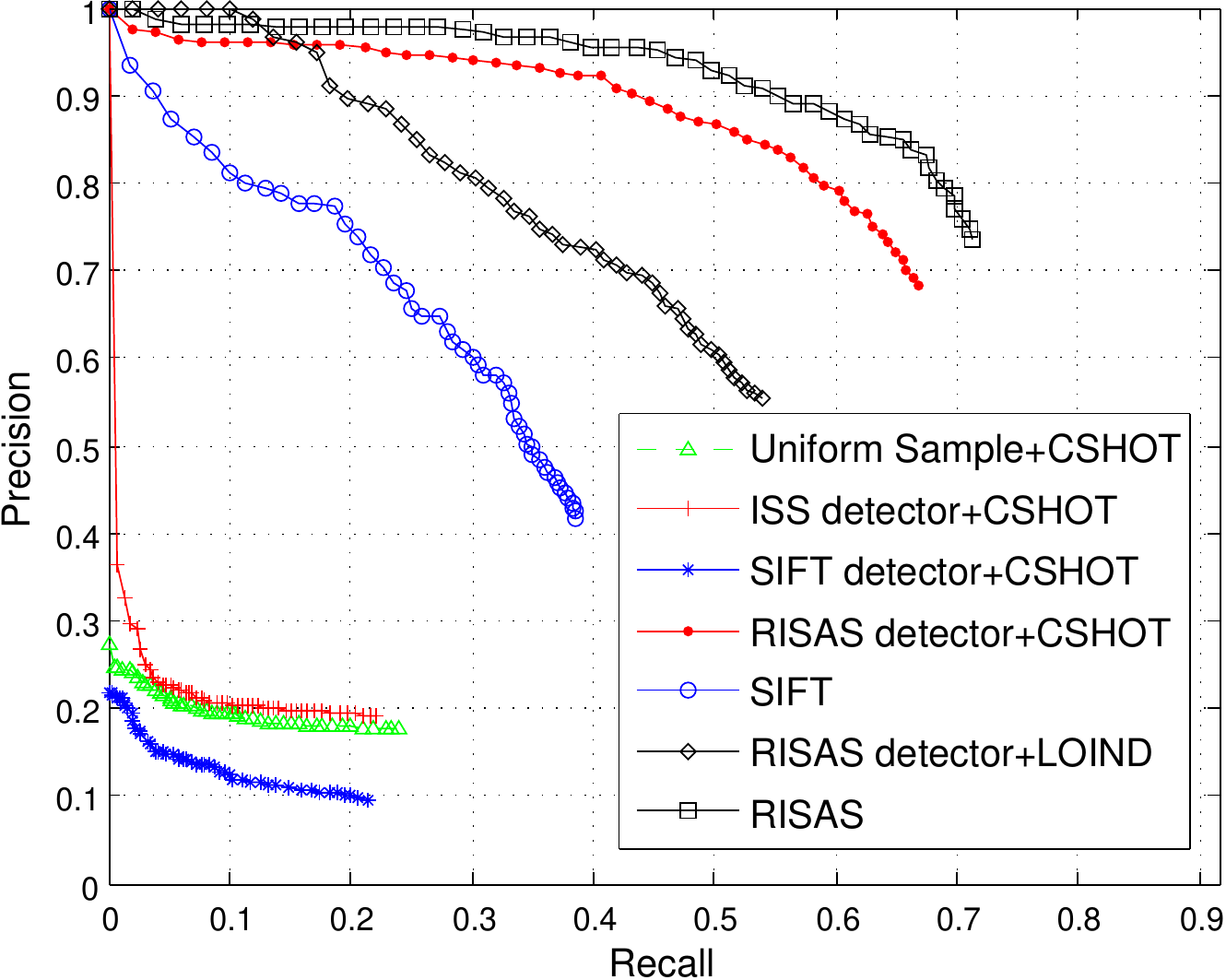}
		\label{fig:vc_pr2}}
	\\
	\subfigure[Between image 12 and 18]
	{\includegraphics[width=0.42\linewidth]{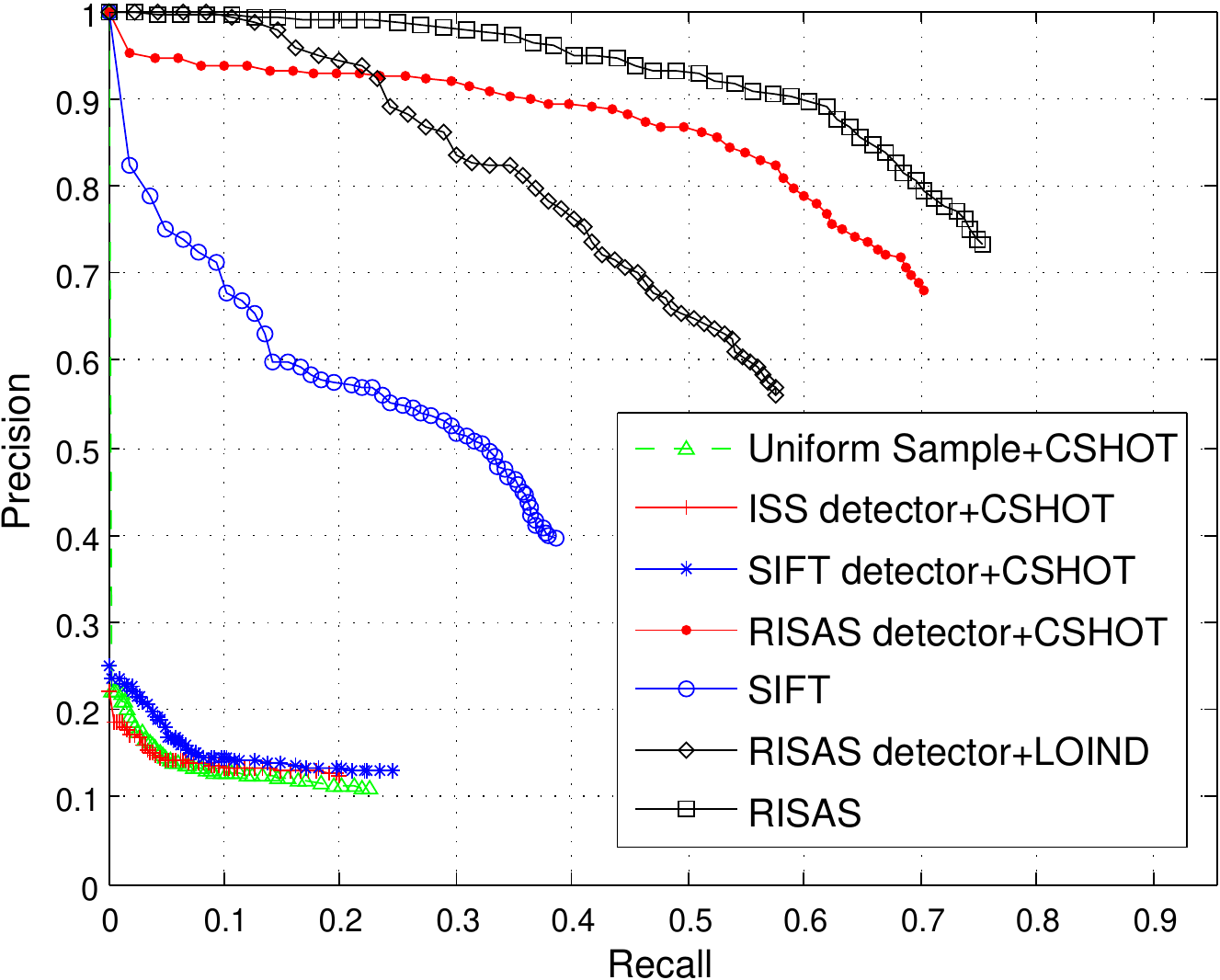}
		\label{fig:vc_pr3}}
	\subfigure[Between image 12 and 24]
	{\includegraphics[width=0.42\linewidth]{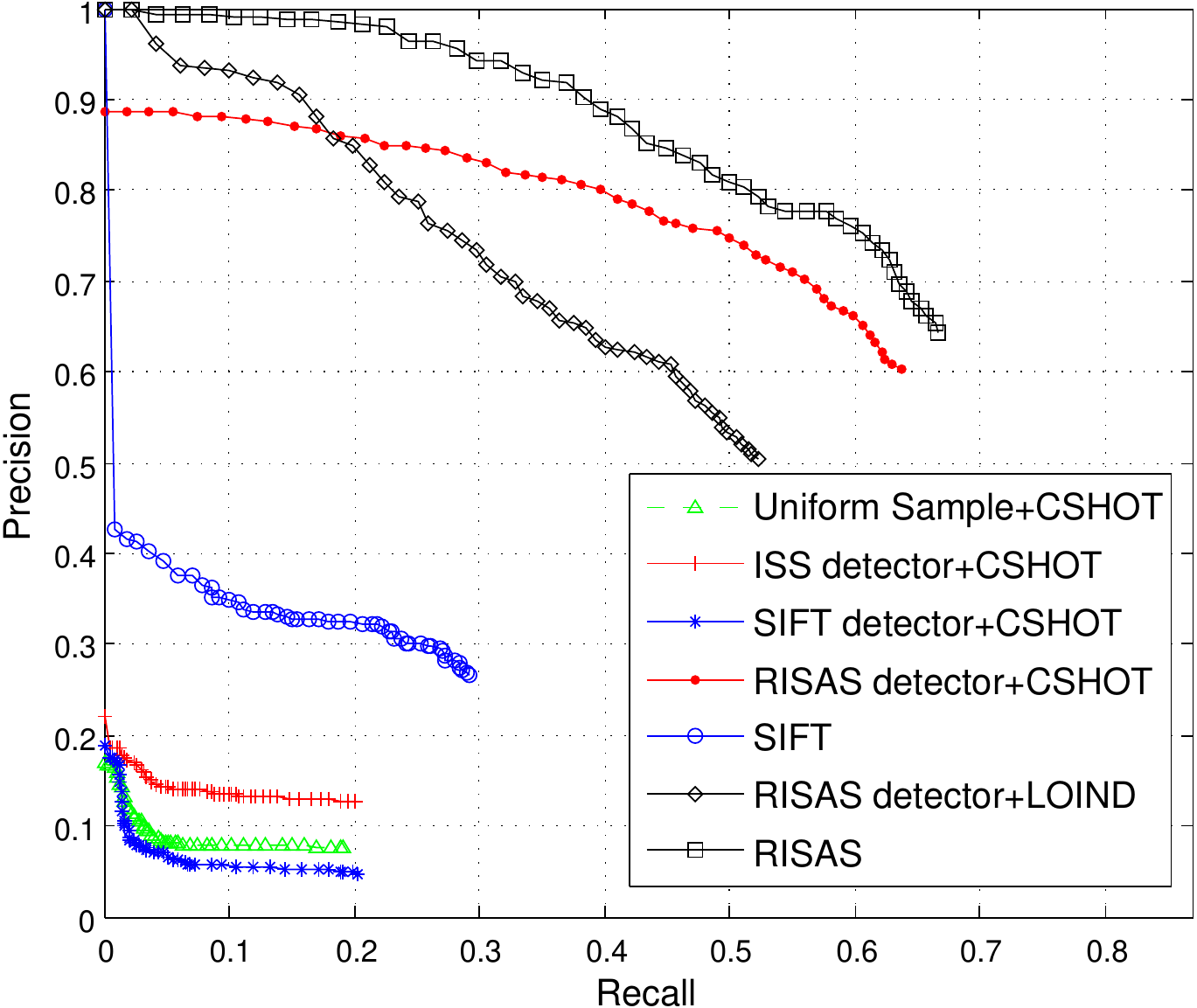}
		\label{fig:vc_pr4}}	
	\caption{Precision-Recall curves under viewpoint variations.}\label{fig:vc_pr}
\end{figure}

We collected $ 24 $ images by moving the sensor around the objects in approximately $ 60^{\circ} $ at $ 0.7 $ meters away from the objects. The angle between each pair of consecutive frames is approximately $ 3^{\circ} $.  In order to estimate the true transformation between each pair of frames and to further evaluate the performance of descriptors, we adopted RGBD-SLAM\cite{endrestro} to compute the optimised poses and regarded the optimised poses as the ground-truth. We selected the image which faces straight forward to the object (in the middle with index $ 12 $ ) as the reference image and matched two images on both left and right side (with indices $ 1, 6, 18 $ and $ 24 $) to the reference one. Image $ 12 $ and $ 24 $ are presented in Fig. \ref{fig:vp_example}. The Precision-Recall curves of these four pairs of images are shown in Fig. \ref{fig:vc_pr}. RISAS is significantly superior compared with all other methods. CSHOT performs well when used with the RISAS detector while performing surprisingly poor with SIFT and ISS detectors, and also with uniform sampling. We also noticed that SIFT doesn't perform as expected under these scenarios with approximate $ 30^\circ $ of viewpoint change.

\subsubsection{Illumination Invariance} 
\begin{figure}[!ht]
	\centering
	\subfigure[Square root illumination]
	{\includegraphics[width=0.42\linewidth]{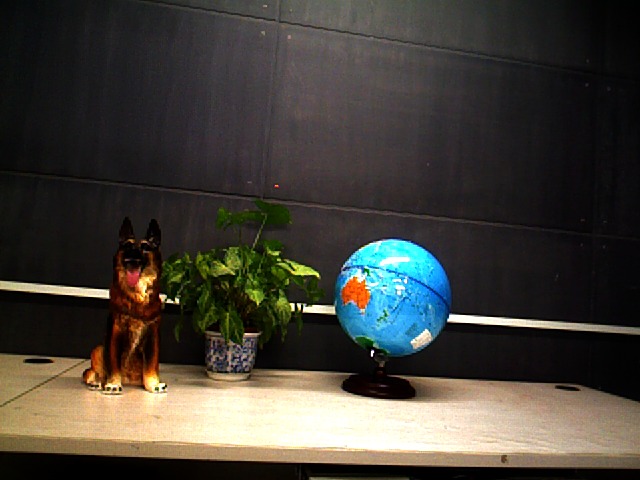}
		\label{fig:illu_pr_nd_img}}
	\subfigure[Precision-Recall curve]
	{\includegraphics[width=0.40\linewidth]{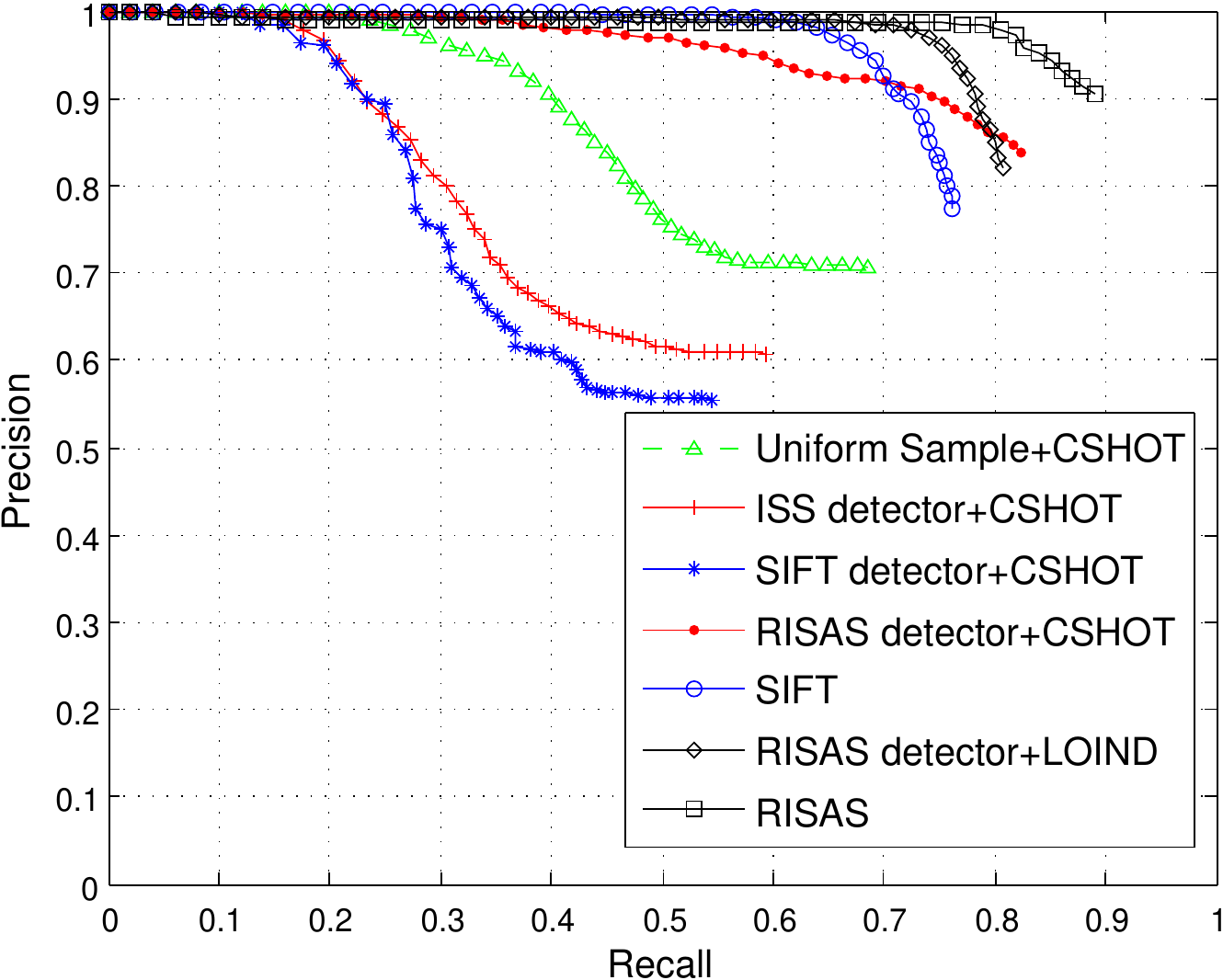}
		\label{fig:illu_pr1}}
	\label{fig:illu_pr_nd}
	\\
	\subfigure[Square illumination]
	{\includegraphics[width=0.42\linewidth]{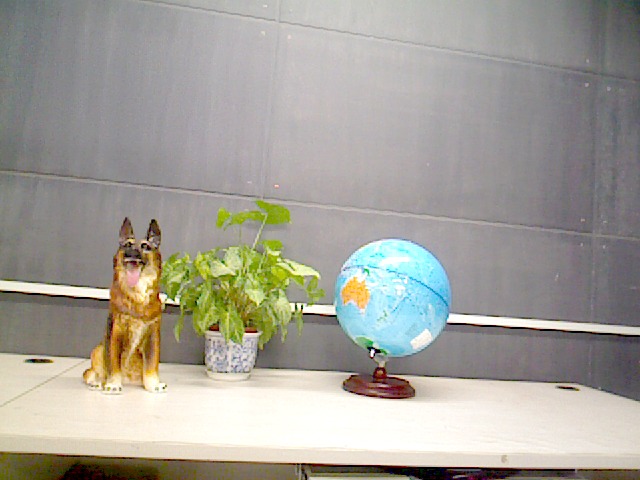}
		\label{fig:illu_pr_nd_img}}
	\subfigure[Precision-Recall curve]
	{\includegraphics[width=0.40\linewidth]{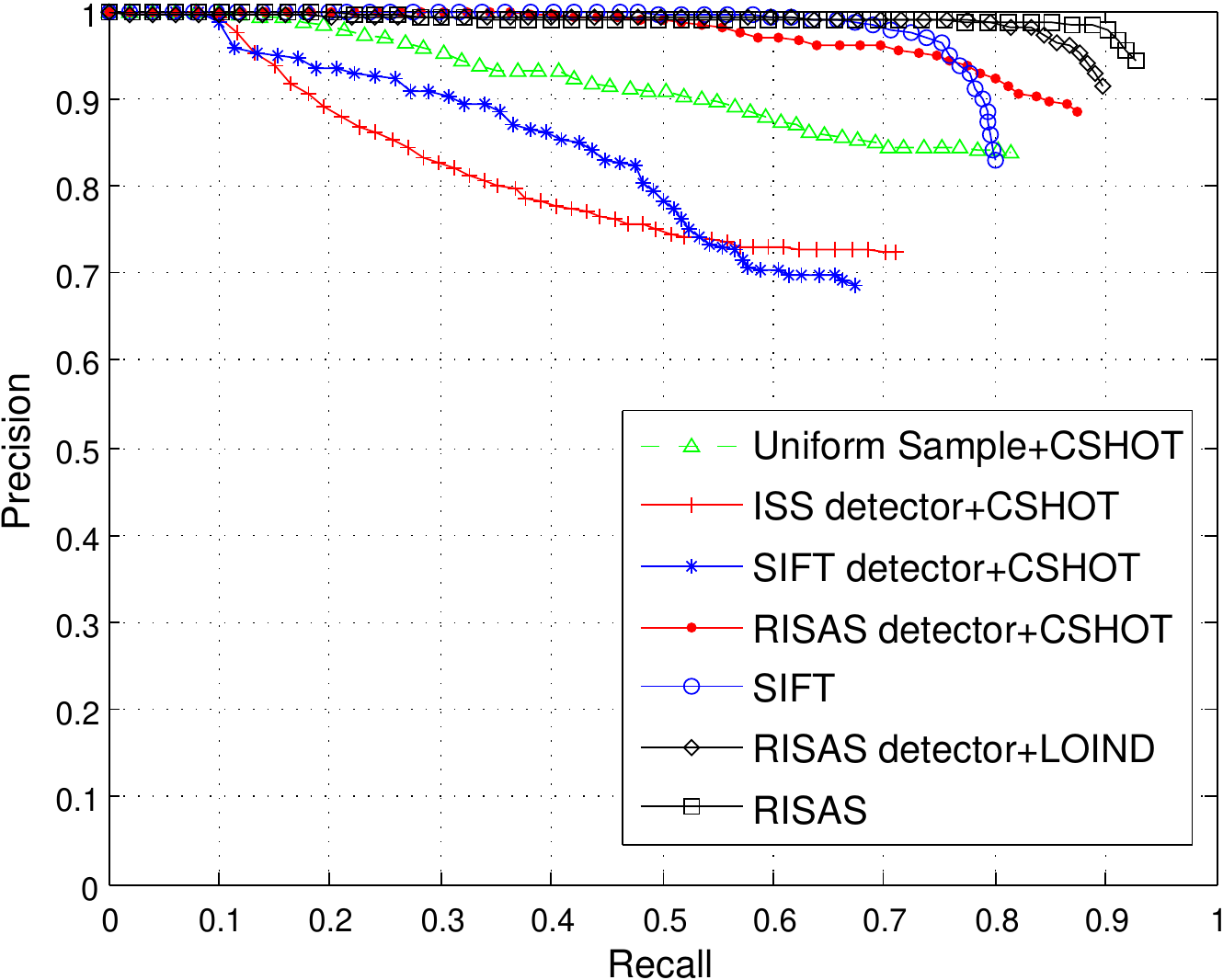}
		\label{fig:illu_pr2}}
	\label{fig:illu_pr_nd}
	\\
	\subfigure[Cube root illumination]
	{\includegraphics[width=0.42\linewidth]{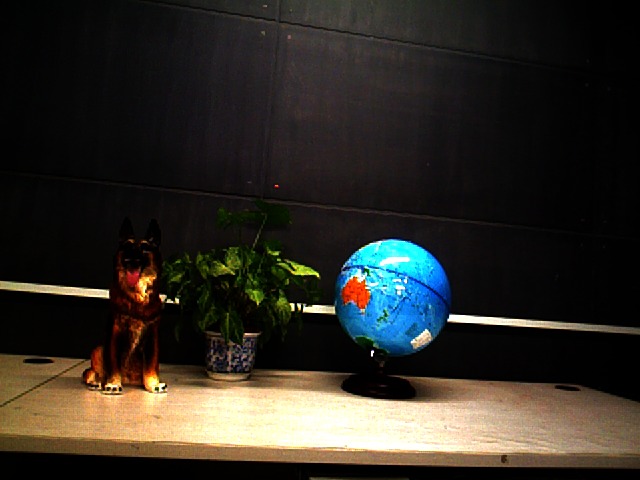}
		\label{fig:illu_pr_nd_img}}
	\subfigure[Precision-Recall curve]
	{\includegraphics[width=0.40\linewidth]{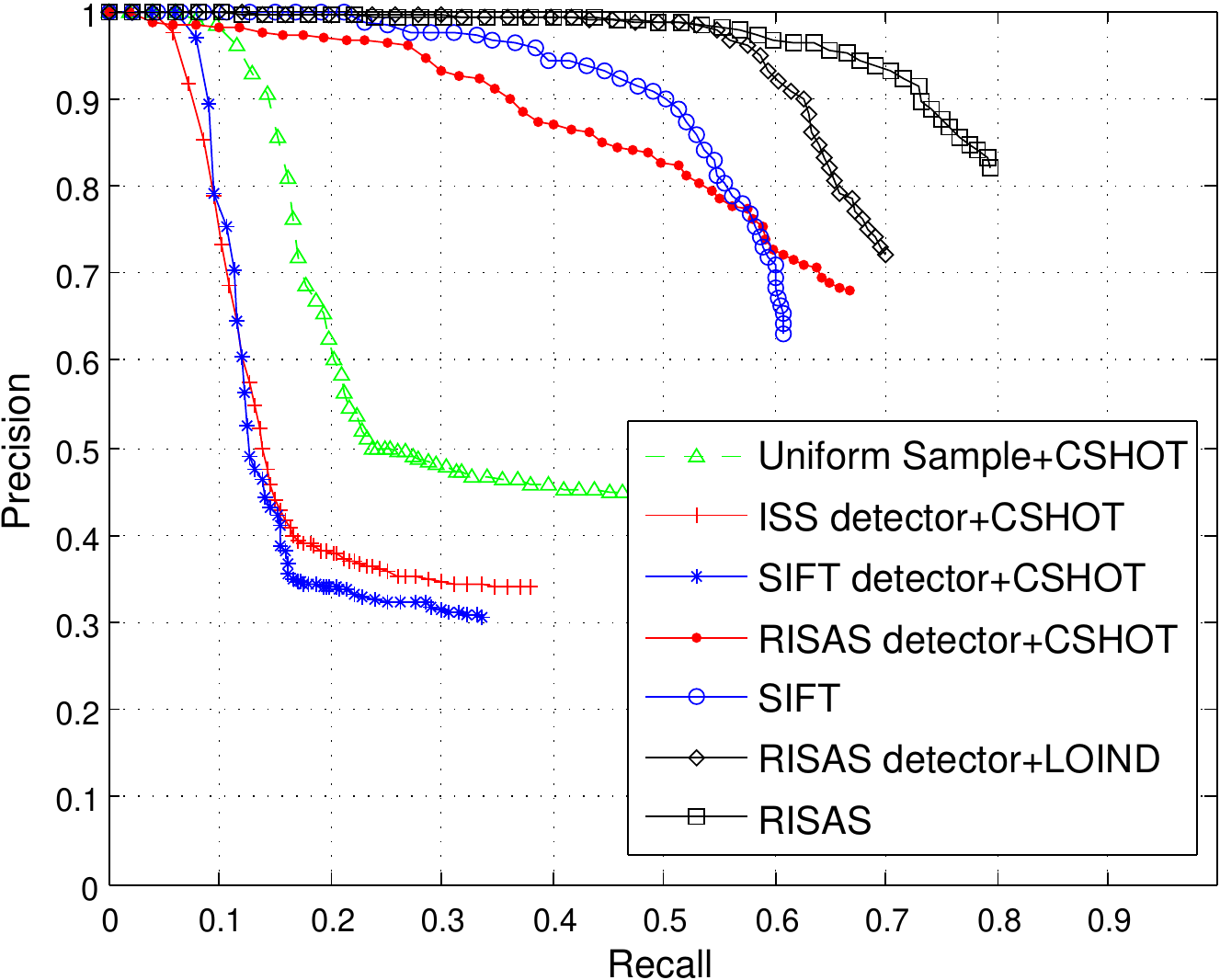}
		\label{fig:illu_pr3}}
	\label{fig:illu_pr_nd}
	\\
	\subfigure[Cube illumination]
	{\includegraphics[width=0.42\linewidth]{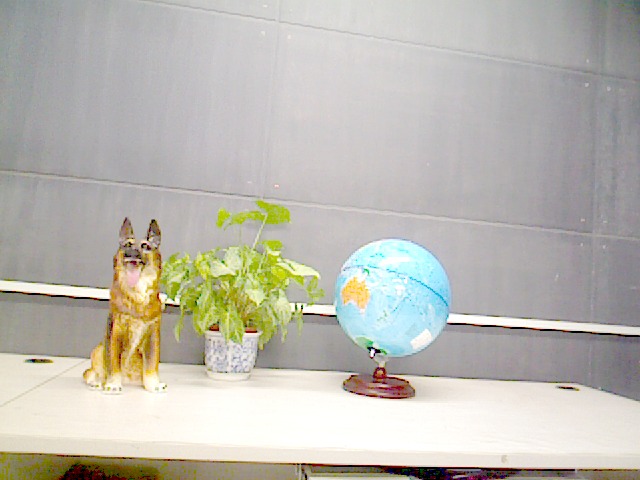}
		\label{fig:illu_pr_nd_img}}
	\subfigure[Precision-Recall curve]
	{\includegraphics[width=0.40\linewidth]{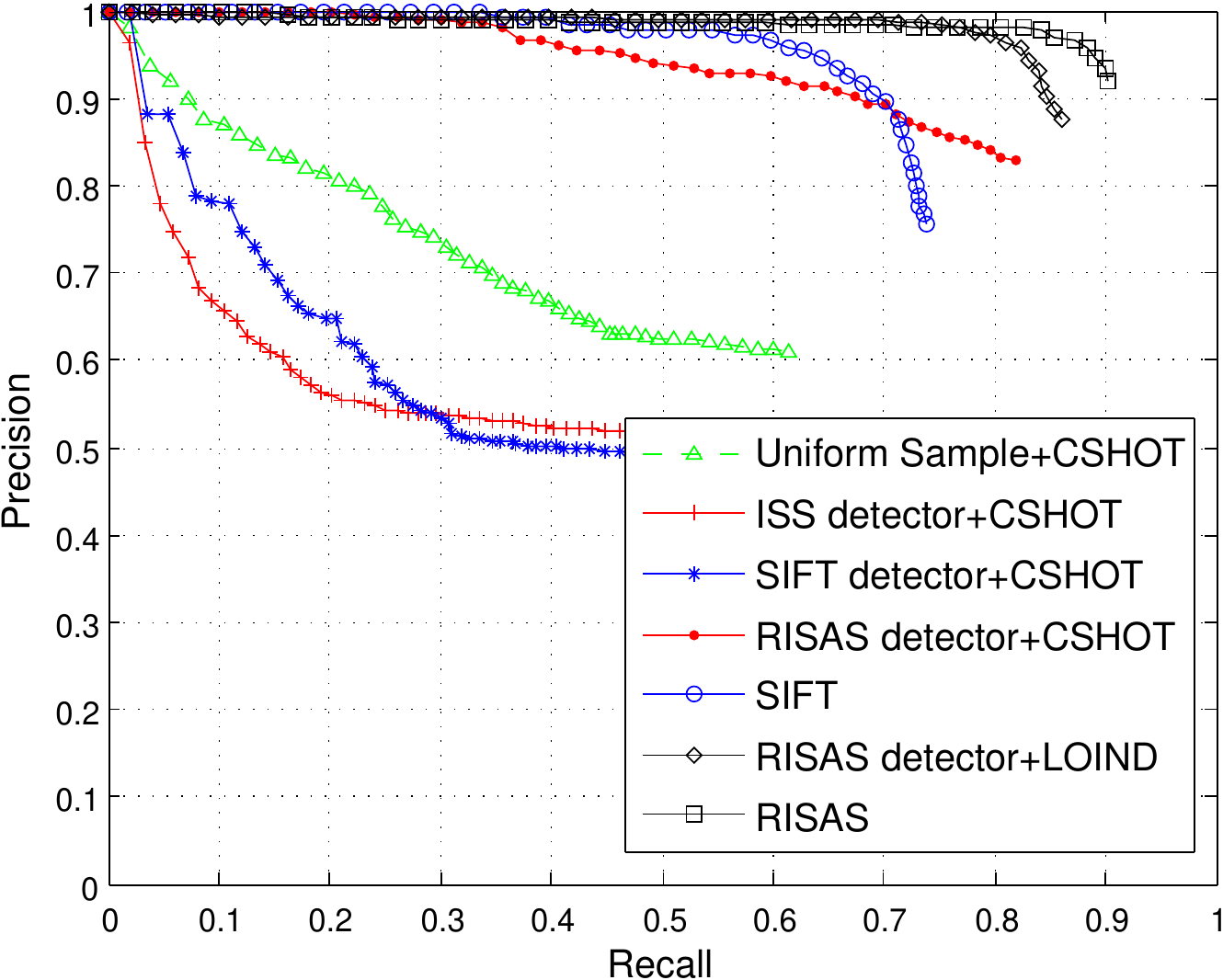}
		\label{fig:illu_pr4}}
	\label{fig:illu_pr_nd}
	\\
	\subfigure[{Illumination change using ND mirror}]
	{\includegraphics[width=0.42\linewidth]{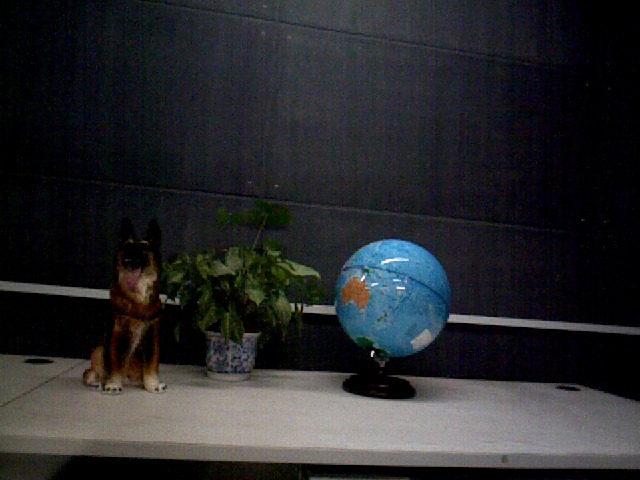}
		\label{fig:illu_pr_nd_img}}
	\subfigure[Precision-Recall curve]
	{\includegraphics[width=0.40\linewidth]{./figures/newresults/illumination_variation_5.pdf}
		\label{fig:illu_pr_nd_pr}}
	\caption{RISAS evaluation under illumination variations.}\label{fig:illu_pr}
\end{figure}

\begin{figure}[htbp]
	\centering
	\centering
	\includegraphics[width=0.45\linewidth]{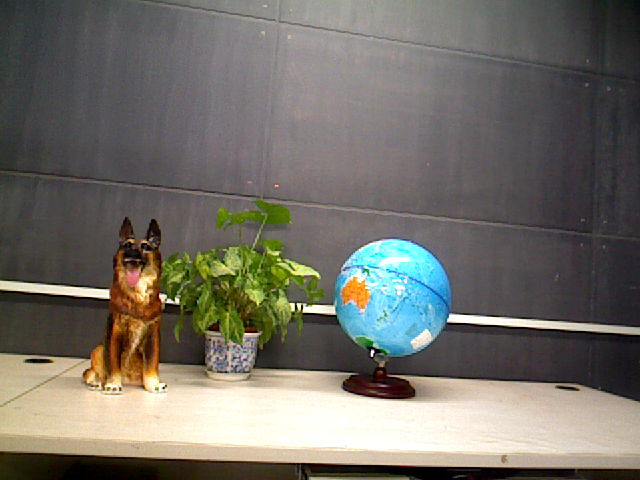}
	\caption{Reference image for illumination and rotation variation.}
	\label{fig:illu_ref}
\end{figure}

In order to validate the performance of RISAS under illumination variations, we constructed a dataset which consists of five different levels of illumination variations: 1) square 2) square root 3) cube, 4) cube root and 5) natural illumination variation, as shown in the left column in Fig. \ref{fig:illu_pr}. The reference image is show in Fig. \ref{fig:illu_ref}. As Fig. \ref{fig:illu_pr} demonstrates, the proposed RISAS feature shows the best performance compared with other approaches, i.e. the precision value of RISAS is almost equal to $ 1.0 $ when the recall value is $0.7$ regardless of the extent of the illumination variation. It is interesting to note that SIFT performs quite well while at the same time performance of CSHOT is significantly enhanced by using it together with the RISAS detector.

\subsubsection{Scale Invariance} 
\begin{figure}[htbp]
	\centering
	\subfigure[Original image as reference, captured at distance $ \approx 1.1m $]
	{\includegraphics[width=0.42\linewidth]{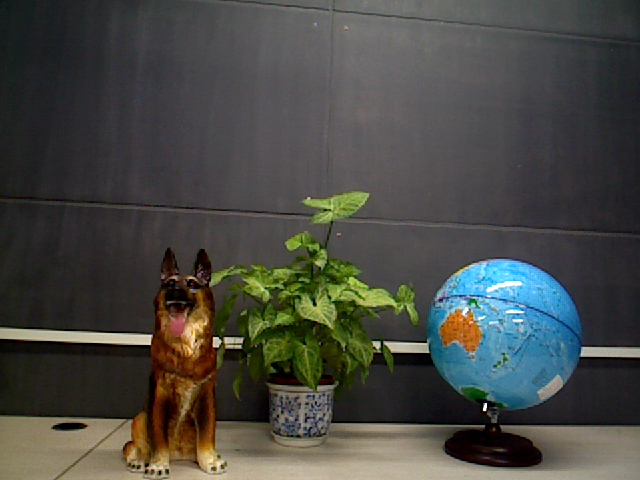}
		\label{fig:scale_1}}
	\subfigure[Image captured at the distance $ \approx 1.9m $]
	{\includegraphics[width=0.42\linewidth]{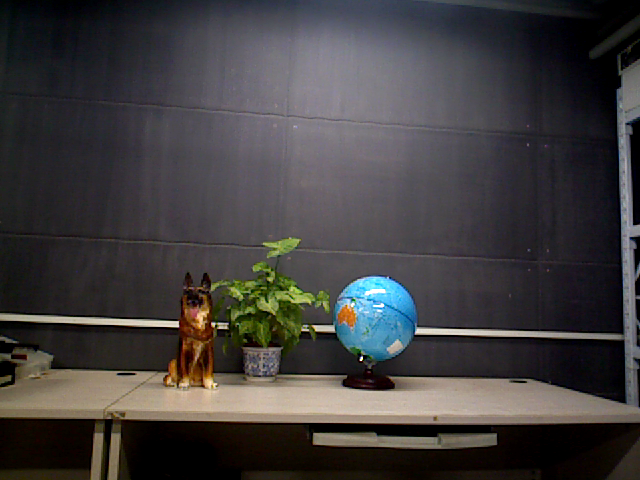}
		\label{fig:scale_10}}
	\caption{Example images of scale variations.}\label{fig:scale_example}
\end{figure}

\begin{figure}[htbp]
\centering
\centering
\includegraphics[width=0.9\linewidth]{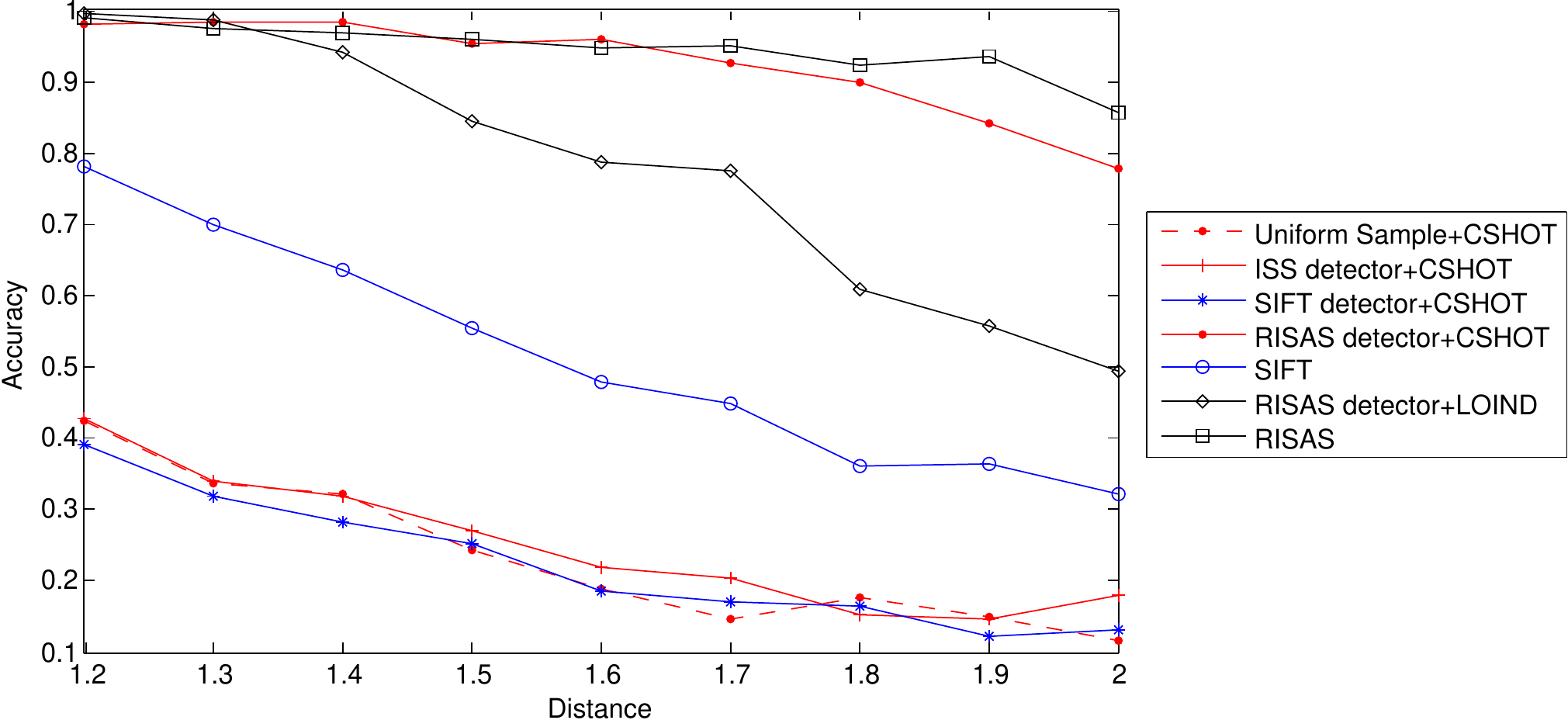}
\caption{Comparative matching results under scale variations.}
\label{fig::scale}
\end{figure}

In this experiment, we collected $ 10 $ images with the variations in $ z $ axis of the sensor coordinate system. The first frame captured at $ 1.1 $ m from the object was selected as the reference image and all other images were captured by moving the camera backwards in step of $ 0.1 $ m. A pair of images of scale variations is shown in Fig. \ref{fig:scale_example} and the matching accuracy w.r.t the scale variation is shown in Fig. \ref{fig::scale}. While RISAS gives the best performance, RISAS detector used with CSHOT also demonstrates good results. All the other methods are significantly inferior.

\subsubsection{Rotation Invariance} 
\begin{figure}[!htbp]
	\centering
	\includegraphics[width=0.9\linewidth]{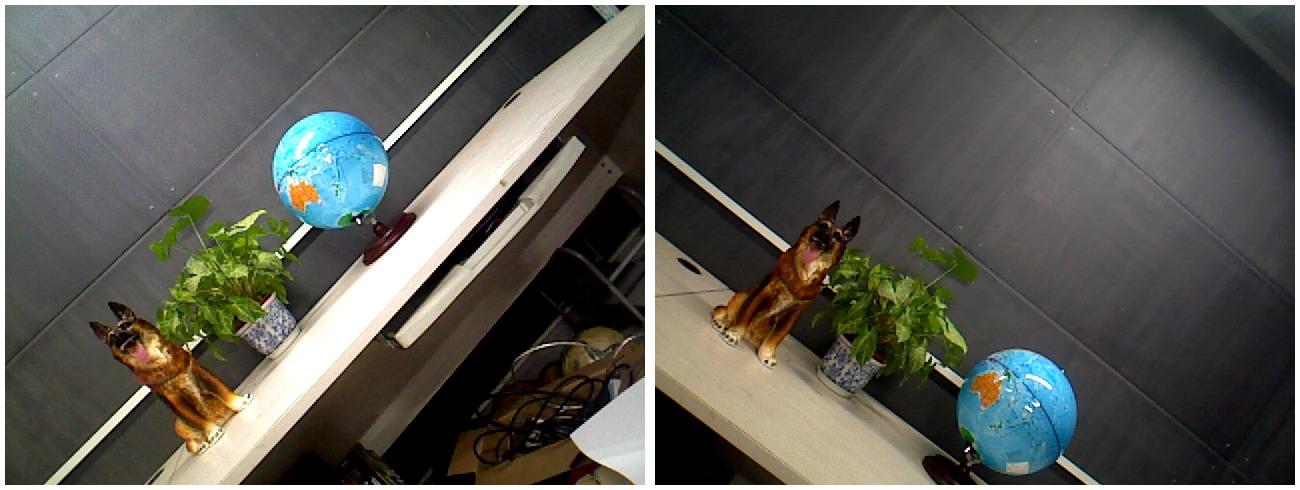}	
	\caption{Example images of 3D rotations.}
	\label{fig::rotate}
\end{figure}

\begin{figure}[!htbp]
	\centering
	\subfigure[]
	{\includegraphics[width=0.45\linewidth]{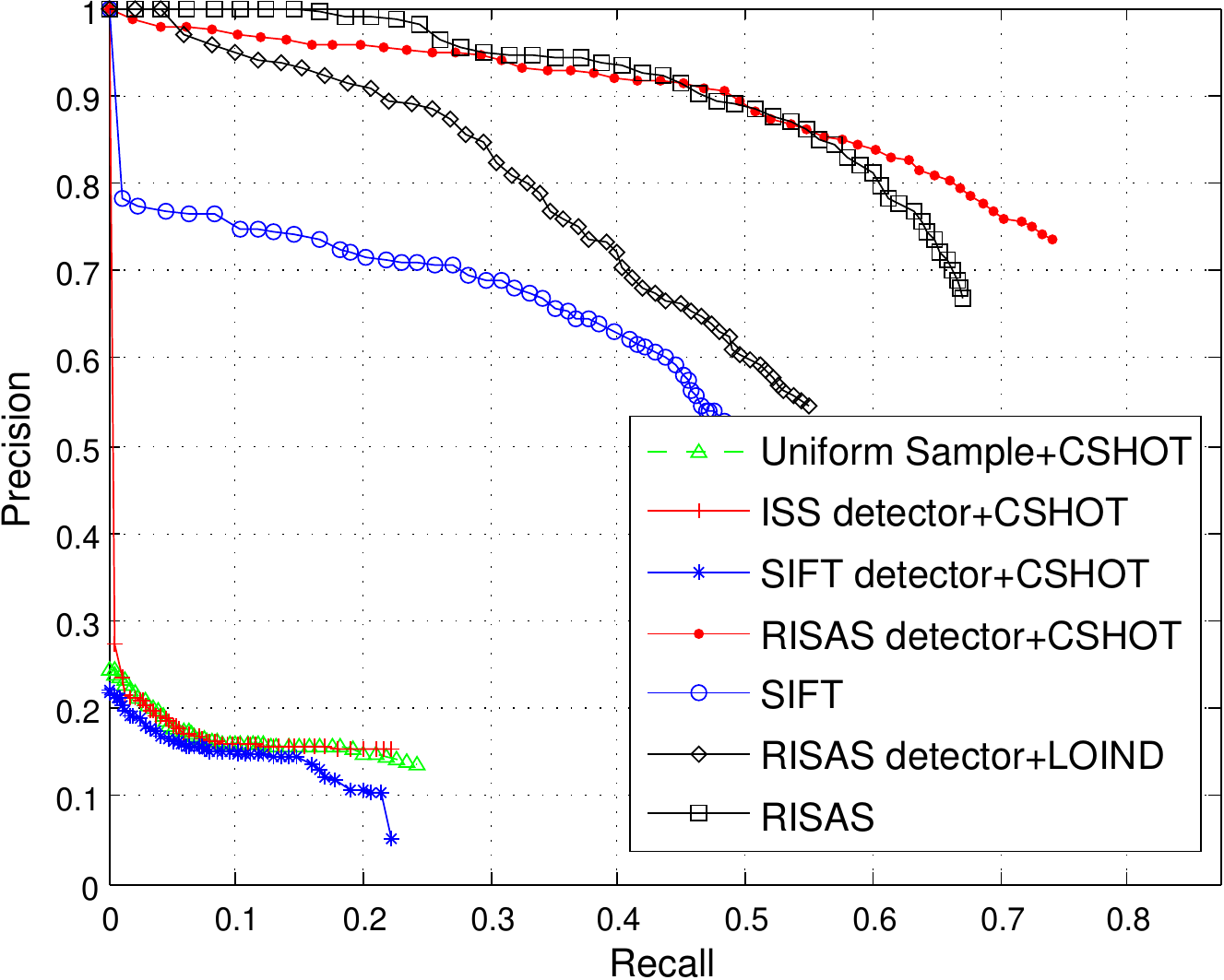}
		\label{fig::rotate_pr1}}1
	\subfigure[]
	{\includegraphics[width=0.45\linewidth]{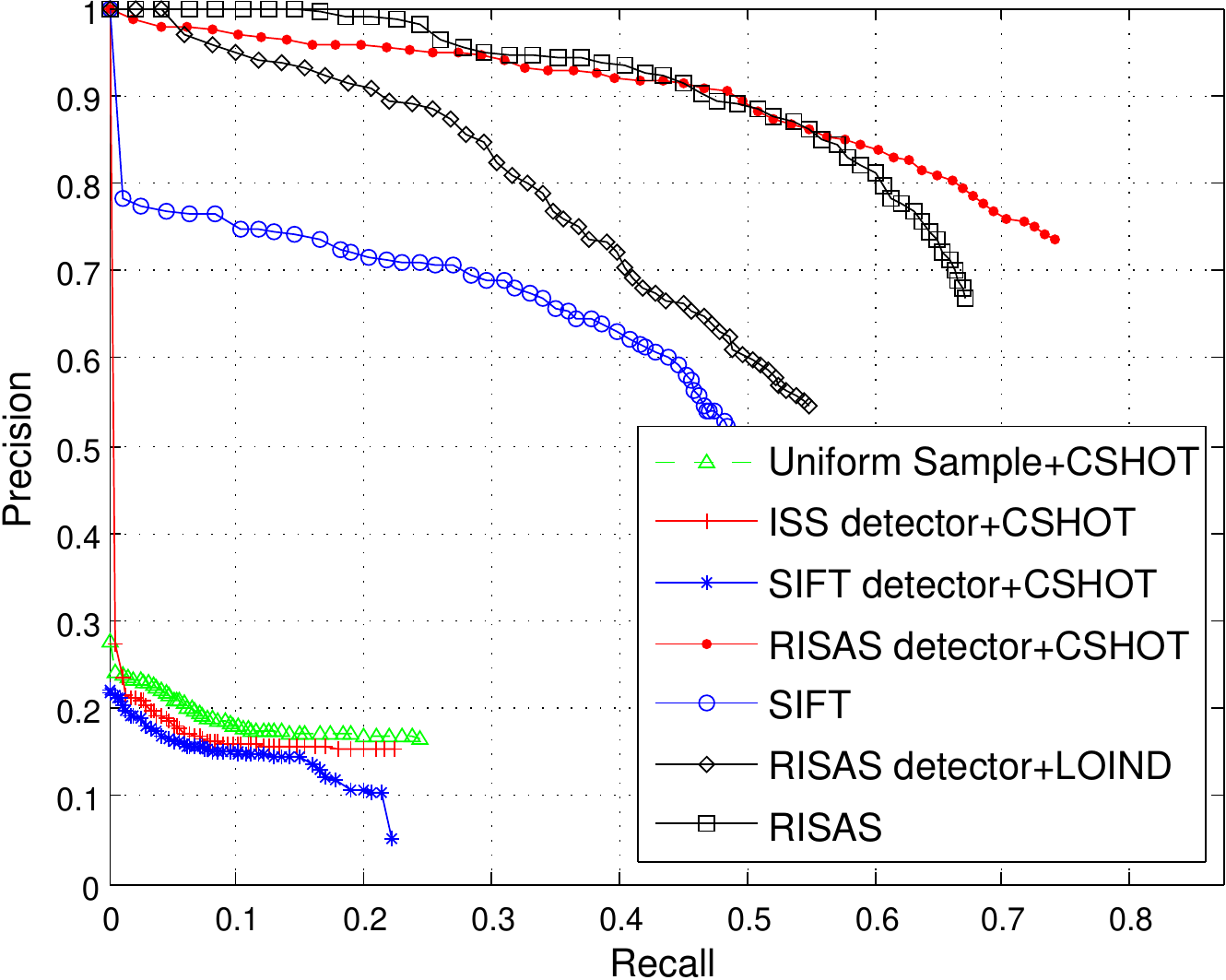}
		\label{fig::rotate_pr2}}
	\caption{Precision-Recall curves corresponding to Fig. \ref{fig::rotate}.}
	\label{fig::rotate_pr}
	\vspace{-2mm}
\end{figure}

We evaluated RISAS under 3D rotation as illustrated in Fig. \ref{fig::rotate}. The reference image is shown in Fig. \ref{fig:illu_ref} for illumination variations. Precision-recall curves are presented in Fig. \ref{fig::rotate_pr}.  RISAS and the combination of RISAS detector and CSHOT performs best under 3D rotations.
	
\subsection*{\textbf{Discussion}} 
Results from the experiments shows that overall, RISAS provides the best results when compared with other approaches. RISAS shows clear advantages over other methods under viewpoint variations. Under illumination variations, RISAS outperforms other methods significantly except for LOIND. For the case of LOIND results are comparable. Under scale and rotation variations, RISAS and the combination of RISAS detector and CSHOT descriptor demonstrate the best performance.

It is clear that using the RISAS detector with CSHOT significantly enhances its matching performance. This confirms our view that a suitable RGB-D detector is critical for the performance of a RGB-D descriptor. In RISAS, the descriptor performs well if the neighbourhood of the keypoint shows higher normal vector variations. This variation is precisely what we consider in developing the detector.

In its current unoptimised Matlab based implementation, RISAS takes  $ 20 $ seconds to complete both keypoint detection and descriptor construction for an image $ 640\times 480 $ captured from Kinect/Xtion. On the same PC with C/C++ implementations in PCL \cite{rusu20113d}, ISS\cite{zhong2009intrinsic} takes nearly 6 seconds and CSHOT takes almost 1 second to process a similar frame. Our expectation is that RISAS can be speeded up to about 2 seconds/frame when implemented in C/C++.

\section{Conclusion}

This paper presents an RGB-D feature which consists of a highly coupled RGB-D keypoint detector and descriptor. A novel 3D representation, dot-product image is combined with grayscale image to extract the keypoints using a principle similar to that of the Harris detector. We also propose an enhanced RGB-D descriptor based on our previous LOIND descriptor which significantly improves the matching performance. RISAS is demonstrated to be invariant to viewpoint, illumination, scale and rotation. RISAS detector is shown to enhance the performance of CSHOT and LOIND that are currently the best performing RGB-D descriptors. Future work will focus on a public release of a C/C++ implementation of RISAS as well as further empirical evaluations.






\bibliographystyle{IEEEtran}
\bibliography{ref}

\end{document}